\RequirePackage{fix-cm}
\documentclass[twocolumn]{svjour3}          
\smartqed  
\usepackage[dvips]{graphicx}
\usepackage{amssymb}
\usepackage{graphicx}
\usepackage{amsmath}
\usepackage{bbm}
\usepackage[sort&compress, round, authoryear]{natbib}
\begin{document}

\title{Optimal Weights Mixed Filter for Removing Mixture of Gaussian
and Impulse Noises
}


\author{Qiyu Jin         \and
        Ion Grama        \and
        Quansheng Liu
}


\institute{Q. Jin \at
               UMR 7590 CNRS, Institut de Min\'{e}ralogie et de Physique des Milieux Condens\'{e}s, Universit\'{e} Pierre et Marie Curie,  Campus Jussieu,
4 place Jussieu,
75005 Paris  \\
              Tel.: +33-144275241\\
              \email{Jin.Qiyu@impmc.upmc.fr}           
           \and
           I. Grama \at
               UMR 6205, Laboratoire de Mathmatiques de Bretagne Atlantique, Universit\'{e} de Bretagne-Sud, Campus de Tohaninic, BP 573,
56017 Vannes, France
\\
Universit\'{e} Europ\'{e}ne de Bretagne, France
\\
              Tel.: +33-297017215\\
              \email{ion.grama@univ-ubs.fr}           
                    \and
           Q. Liu \at
               UMR 6205, Laboratoire de Mathmatiques de Bretagne Atlantique, Universit\'{e} de Bretagne-Sud, Campus de Tohaninic, BP 573,
56017 Vannes, France
\\
Universit\'{e} Europ\'{e}ne de Bretagne, France
\\
              Tel.: +33-297017140\\
              \email{quansheng.liu@univ-ubs.fr}
}

\date{Received: date / Accepted: date}

\maketitle

\begin{abstract}
According to the character of Gaussian, we modify the Rank-Ordered Absolute Differences (ROAD) to Rank-Ordered Absolute Differences of mixture of
Gaussian and impulse noises (ROADG). It will be more effective to detect impulse noise
when the impulse is mixed with Gaussian noise. Combining rightly the ROADG
with Optimal Weights Filter (OWF), we obtain a new method to deal with
the mixed noise, called Optimal Weights Mixed Filter (OWMF). The
simulation results show that the method is effective to remove the mixed
noise.
\end{abstract}

\section{Introduction\label{sec introduction}}

Noise can be systematically introduced into digitized images during
acquisition and transmission, which usually degrade the quality of
digitized images. However, various image-related applications, such as
aerospace, medical image analysis, object detection etc., generally require
effective noise suppression to produce reliable results. The problem of
noise removal from a digitized image is one of the most important ones in
digital image processing. The nature of the problem depends on the type of
noise to the image. Generally, two noise models can adequately represent
most noise added to images. Often in practice it is assumed that the noise has
two components: an additive Gaussian noise and an impulse noise.

The additive Gaussian noise model is:
\begin{equation}
Y(x)=f(x)+\epsilon (x),\;x\in \mathbf{I,}  \label{Gaussian model}
\end{equation}%
where $\mathbf{I}=\{\frac{1}{N},\frac{2}{N},\cdots ,\frac{N-1}{N},1\}^{2},$ $%
N\in \mathbf{N}$, $Y$ is the observed image brightness, $f:\mathbf{I}\mapsto
\lbrack c,d]$ is an unknown target regression function, and $\epsilon (x),$ $%
x\in \mathbf{I,}$ are independent and identically distributed (i.i.d.)
Gaussian random variables with mean $0$ and standard deviation $\sigma>0.$
The additive Gaussian noise is characterized by adding to each digitized
image pixel a value from a zero-mean Gaussian distribution. Such noise is
usually introduced during image acquisition. The zero-mean property of this
Gaussian distribution makes it possible to remove the Gaussian noise by
Non-Local weighted averaging. Important denoising methods for the Gaussian
noise model have been well developed in recent years, see for example
 \citet{polzehl2000adaptive},
 \citet{buades2005non},
\citet{kervrann2006optimal},
\citet{aharon2006rm},
\citet{dabov2007image},
\citet{cai2008two},
 \citet{hammond2008image},
 \citet{lou2010image},
\citet{Katkovnik2010local},  and
\citet{jin2011removing}.

The random impulse noise model is: for $x\in \mathbf{I,}$
\begin{equation}
Y\left( x\right) =\left\{
\begin{array}{ll}
n(x), & \text{if }x\in \mathbf{B}, \\
f(x), & \text{if }x\in \mathbf{I}\backslash \mathbf{B},%
\end{array}
\right.  \label{Impulse model}
\end{equation}%
where $\mathbf{B}$ is the set of pixels contaminated by impulse noise, $\mathbb{P}(%
\mathbf{B})=p$ is the impulse probability (the proportion of the occurrence
of the impulse noise), $n(x)$ are independent random variables uniformly
distributed on some interval $[c,d]$. The impulse noise is characterized by
replacing a portion of an image's pixel values with random values, leaving
the remaining one unchanged. Such a noise can be introduced due to
transmission errors, malfunctioning pixel elements in the camera sensors,
faulty memory locations, and timing errors in analog-to-digital conversion.
Recently, some important methods  have been proposed to remove the impulse
noise, see for example:
\citet{hwang1995adaptive},
\citet{abreu1996new},
 \citet{chen2001space},
 \citet{chan2004iterative},
 \citet{nikolova2004variational},
\citet{wenbin2005new},
\citet{dong2007detection}, and
 \citet{yu2008efficient},

To remove a mixture of the Gaussian and impulse noises which defined by
\begin{equation}
Y\left( x\right) =\left\{
\begin{array}{ll}
n(x), & \text{if }x\in \mathbf{B}, \\
f(x)+\epsilon (x), & \text{if }x\in \mathbf{I}\backslash \mathbf{B},%
\end{array}%
\right.  \label{mixture model}
\end{equation}
the above mentioned
methods are not effective: the Gaussian noise removal methods cannot
adequately remove impulse noise, for they interpret the impulse noise pixel
as edges to be preserved; when impulse removal methods are applied to an
image corrupted with the Gaussian noise, such filters, in practice, leave
grainy, visually disappointing results. \citet{garnett2005universal} introduced a new local image statistic called
Rank Ordered Absolute Difference (ROAD) to identify the impulse noisy pixels
and incorporated it into a filter designed to remove the additive Gaussian
noise. As a result they have obtained a trilateral filter capable to remove mixed
Gaussian and impulse noise. This method also performs well for removing the
single impulse noise. For other developpements in this direction we refer to
\citet{bouboulis2010adaptive},
\citet{li2010new},
\citet{xiao2011restoration}),
\citet{luisier2011image}.

In this paper,
we propose a new filter that we call \textit{Optimal Weights Mixed
Filter} (OWMF). The idea of our method come from the  combination of the ROAD statistic of %
\citet{garnett2005universal} and  the Optimal Weights Filter in %
\citet{jin2011removing}. In the paper, we modify the Rank-Ordered Absolute Differences (ROAD) to Rank-Ordered Absolute Differences of mixture of
Gaussian and impulse noises (ROADG). It will be more effective to detect impulse noise
when the impulse is mixed with Gaussian noise.
The ROADG statistic will give a weight for all pixels in the image, which take value in the interval $(0,1]$. The weight will get low value (the value may be near to 0) when a pixel is contaminated by impulse noise; otherwise, the weight will carry a high value (the value may be near to 1). Then the  Optimal Weights Filter (OWF) combining with the  ROADG statistic can detect the impulse points in the image and give the proper weights  to deal with the mixed noise.  As a result, we obtain our new filter. The simulation results show that the proposed
filter can effectively remove the mixture of impulse noise and the Gaussian
noise. Moreover, when applied to either the single impulse noise or the
single Gaussian noise it performs as good as the best filters specialized to
single noises.

The rest of the paper is organized as follows. In Section \ref{sec algorithms}
after a short recall of the Optimal Weights Filter and a brief presentation
of the Trilateral Filter  whose main ideas will be used in the definition of our new filter,
we introduce our filter. In section \ref{sec Simulation and comparisons},
we provide visual examples and numerical results that demonstrate our method's soundness.
Section \ref{sec conclusion} is a brief conclusion.

\section{Algorithms\label{sec algorithms}}

\subsection{Optimal Weights Filter\label{sec Optimal Weights
Filter}}

For any pixel $x_{0}\in \mathbf{I}$ and a given $h>0,$ the square window of
pixels
\begin{equation}
\mathbf{U}_{x_{0},h}=\left\{ x\in \mathbf{I}:\Vert x-x_{0}\Vert _{\infty
}\leq h\right\}  \label{def search window}
\end{equation}%
will be called \emph{search window} at $x_{0}$, where $h$ is a positive integer.
 The size of the square search window $\mathbf{U}_{x_{0},h}$ is the
positive integer number $M=(2h+1)^{2}=\mathrm{card\ }\mathbf{U}_{x_{0},h}.$ For
any pixel $x\in \mathbf{U}_{x_{0},h}$ and a given integer $\eta >0$ a second square
window of pixels
\begin{equation}
\mathbf{V}_{x,\eta }=\left\{ y\in \mathbf{I}:\Vert y-x\Vert _{\infty
}\leq \eta \right\}  \label{def patch}
\end{equation}%
will be called for short a \emph{patch window} at $x$ in order to be
distinguished from the search window $\mathbf{U}_{x_{0},h}.$  The size of
the patch window $\mathbf{V}_{x,\eta }$ is the positive integer $m=(2\eta+1)
^{2}=\mathrm{card\ }\mathbf{V}_{x_{0},\eta }.$ The vector $\mathbf{Y}%
_{x,\eta }=\left( Y\left( y\right) \right) _{y\in \mathbf{V}_{x,\eta }}$
formed by the values of the observed noisy image $Y$ at pixels in the
patch $\mathbf{V}_{x,\eta }$ will be called simply \emph{data patch} at $%
x\in \mathbf{U}_{x_{0},h}.$ For any $x_{0}\in \mathbf{I}$ and any $x\in
\mathbf{U}_{x_{0},h}$,  a distance between the data patches $\mathbf{Y%
}_{x,\eta }=\left( Y\left( y\right) \right) _{y\in \mathbf{V}_{x,\eta }}$
and $\mathbf{Y}_{x_{0},\eta }=\left( Y\left( y\right) \right) _{y\in \mathbf{%
V}_{x_{0},\eta }}$ is defined by
\begin{equation}
\mathbf{d}^{2}\left( \mathbf{Y}_{x,\eta },\mathbf{Y}_{x_{0},\eta }\right) =%
\frac{1}{m}\left\Vert \mathbf{Y}_{x,\eta }-\mathbf{Y}_{x_{0},\eta
}\right\Vert _{2}^{2},  \label{distance l2}
\end{equation}%
where $$\left\Vert \mathbf{Y}%
_{x,\eta }-\mathbf{Y}_{x_{0},\eta }\right\Vert
_{2}^{2}=\sum\limits_{y\in \mathbf{V}_{x_{0},\eta }}(
Y(T_x y)-Y(y)) ^{2}$$
and $T_x$ is the translation mapping: $T_x y=x+(y-x_{0})$.
As
\begin{equation*}
Y(T_x y)-Y(y)=f(T_x y)-f(y)+\epsilon (T_x y)-\epsilon (y)
\end{equation*}%
we have
\begin{equation*}
\mathbb{E}(Y(T_x y)-Y(y))^{2}=(f(T_x y)-f(y))^{2}+2\sigma ^{2}.
\end{equation*}%
If we use the approximation
\begin{equation*}
(f(T_x y)-f(y))^{2}\approx (f(x)-f(x_{0}))^{2}=\rho _{f,x_{0}}^{2}(x)
\end{equation*}%
and the law of large numbers, it seems reasonable that
\begin{equation*}
\rho _{f,x_{0}}^{2}(x)\approx \mathbf{d}^{2}(\mathbf{Y}_{x,\eta }-\mathbf{Y}%
_{x_{0},\eta })-2\sigma ^{2}.
\end{equation*}%
As shown in \citep{jin2011removing}, in practice, a much better denoising
results are obtained by using the following approximation
\begin{equation}
\rho _{f,x_{0}}(x)\approx \widehat{\rho }_{x_{0}}(x)=\left( d(\mathbf{Y}%
_{x,\eta }-\mathbf{Y}_{x_{0},\eta })-\sqrt{2}\sigma \right) ^{+}.
\label{rho-estim}
\end{equation}%
The fact that $\widehat{\rho }_{x_{0}}(x)$ is a good estimator of $\rho
_{f,x_{0}}$ was justified by the convergence results in %
\citep{jin2011removing} (cf. Theorems 3 and 4 of \citep{jin2011removing}). The
Optimal Weights Filter is defined by%
\begin{equation}
\mbox {OWF}(f)(x_{0})=\frac{\sum\limits_{x\in \mathbf{U}_{x_{0},h}}\kappa_{\text{tr}}(%
\frac{\widehat{\rho }_{x_{0}}\left( x\right) }{\widehat{a}})Y(x)}{%
\sum\limits_{y\in \mathbf{U}_{x_{0},h}}\kappa_{\text{tr}}(\frac{\widehat{\rho }%
_{x_{0}}\left( x\right) }{\widehat{a}})},  \label{OWFilter}
\end{equation}%
where $\kappa_{\text{tr}}$ is the usual triangular kernel:
\begin{equation}
\kappa_{\text{tr}}\left( t\right) =\left( 1-\left\vert t\right\vert \right)
^{+},\quad t\in \mathbf{R}^{1}.  \label{def kernel}
\end{equation}

\begin{remark}
\label{calculate a} The bandwidth $\widehat{a}>0$ is the solution of
\begin{equation*}
\sum_{x\in \mathbf{U}_{x_{0},h}}\widehat{\rho }_{x_{0}}(x)(\widehat{a}-%
\widehat{\rho }_{x_{0}}(x))^{+}=\sigma ^{2},
\end{equation*}%
and can be calculated as follows. We sort the set $\{\widehat{\rho }%
_{x_{0}}(x) : x\in \mathbf{U}_{x_{0},h}\}$ in the ascending order $0=\widehat{\rho}_{x_0}(x_1)\leq\widehat{\rho}
_{x_0}(x_2)\leq \cdots \leq \widehat{\rho}_{x_0}(x_M)<\widehat{\rho }_{x_0}(x_{M+1})=+\infty $, where $M=\mathrm{card}\,\mathbf{U}_{x_{0},h}$.
Let
\begin{equation}
a_{k}=\frac{\sigma ^{2}+\sum\limits_{i=1}^{k}\widehat{\rho }_{x_0}(x_{i})^{2}}{%
\sum\limits_{i=1}^{k}\widehat{\rho }_{x_0}(x_{i})},\quad 1\leq k\leq M,  \label{a k}
\end{equation}%
and
\begin{eqnarray}
k^{\ast } &=&\max \{1\leq k\leq M\,:\,a_{k}\geq \widehat{\rho }_{x_0}(x_{k})\}  \notag
\\
&=&\min \{1\leq k\leq M\,:\,a_{k}<\widehat{\rho }_{x_0}(x_{k})\}-1,  \label{k star}
\end{eqnarray}%
with the convention that $a_{k}=\infty $ if $\widehat{\rho }_{x_0}(x_{k})=0$ and that
$\min \varnothing =M+1$. The solution can be expressed as $\widehat{a}%
=a_{k^{\ast }}$; moreover, $k^{\ast }$ is the unique integer $k\in
\{1,\cdots ,M\}$ such that $a_{k}\geq \widehat{\rho }_{x_0}(x_{k})$ and $a_{k+1}<%
\widehat{\rho }_{x_0}(x_{k+1})$ if $k<M$.
\end{remark}

The proof of Remark \ref{calculate a} can be found in \citep{jin2011removing}.

\subsection{ROAD statistic and Trilateral Filter\label{sec ROAD statistic
and Trilateral Filter}}

In \citep{garnett2005universal},  Garnett et al introduced the
Rank-Ordered Absolute Differences (ROAD) statistic to detect points
contaminated by impulse noise. For any pixel $x_{0}\in \mathbf{I}$ and a
given $d>0,$ we define the square window of pixels
\begin{equation*}
\Omega _{x_{0},d}^{0}=\{x:0<N\Vert x-x_{0}\Vert _{\infty }\leq d\},
\end{equation*}%
where $d$ is a positive integer. The square window will be called deleted
neighborhood at $x_{0}$. The ROAD statistic is defined by%
\begin{equation}
ROAD(x_{0})=\sum_{i=1}^{K}r_{i}(x_{0}),\text{ }x_{0}\in \mathbf{I},
\label{Road statistic}
\end{equation}%
where $r_{i}(x_{0})$ is the $i$-th smallest term in the set $%
\{|Y(x)-Y(x_{0})|:x\in \Omega _{x_{0},d}^{0}\}$ and $2\leq K<\mathrm{%
\mathrm{card}\ \Omega _{x_{0},d}^{0}}$. In \citep{garnett2005universal} it is advised
to use $d=1$ and $K=4$. Note that if $x_{0}$ is an impulse noisy point,
the value of $ROAD(x_{0})$ is large; otherwise it is small.

Following \citet{garnett2005universal} and \citet{li2010new} the authors define the
"joint impulsively" $J_{I}\left( x_{0},x\right) $ between $x_{0}$ and $x$
as:
\begin{equation}
J_{I}\left( x_{0},x\right) =\exp \left( -\frac{(ROAD(x_{0})+ROAD(x))^{2}}{%
2(2\sigma _{J})^{2}}\right) ,
\end{equation}%
where the function $J_{I}(x_{0},x)$ assumes values in $[0,1]$ and the
parameter $\sigma _{J}$ controls the shape of the function $J_{I}(x_{0},x).$
If $x_{0}$ or $x$ is an impulse noisy point, then the value of $ROAD(x_{0})$
or $ROAD(x)$ is large and $J_{I}(x_{0},x)\approx 0;$ otherwise, the value of
$ROAD(x_{0})$ and $ROAD(x)$ are small and $J_{I}(x_{0},x)\approx 1.$ The
definition of the trilateral filter (cf. \citep{garnett2005universal}) is given by
\begin{equation*}
\mbox {TriF}(v)(x_{0})=\frac{\sum_{x\in \mathbf{U}_{x_{0},h}}w(x)Y(x)}{%
\sum_{x\in \mathbf{U}_{x_{0},h}}w(x)},
\end{equation*}%
where
\begin{equation*}
\begin{array}{ll}
w(x) & =w_{S}(x)w_{R}(x)^{J_{I}(x_{0},x)}w_{I}(x)^{1-J_{I}(x_{0},x)}, \\
w_{S}(x) & =e^{-\frac{|x-x_{0}|^{2}}{2\sigma _{S}^{2}}}, \\
w_{R}(x) & =e^{-\frac{(Y(x)-Y(x_{0}))^{2}}{2\sigma _{R}^{2}}}, \\
w_{I}(x) & =e^{-\frac{ROAD(x)^{2}}{2\sigma _{I}^{2}}}.%
\end{array}%
\end{equation*}%
This filter has been shown to be very efficient in removing a mixed noise
composed of a Gaussian and random impulse noise.

\subsection{Optimal Weights Mixed Filter \label{sec Optimal Weights Mixed Filter}}

The ROAD statistic (cf. \citet{garnett2005universal}) provides a effective
measure to detection the pixel contaminated by impulse. In this paper, we take
into account the character of Gaussian noise, and modify the method ROAD to
adapt to the mixture of impulse and Gaussian noises. Then the equation (\ref%
{Road statistic}) becomes
\begin{equation}
ROADG(x_{0})=\left( \frac{1}{K}\sum_{i=1}^{K}r_{i}(x_{0})-\sigma \right)
^{+},\;x_{0}\in \mathbf{I},
\end{equation}%
where $\sigma $ is the standard deviation of the added Gaussian noise, $r_{i}(x_{0})$
is the $i$-th smallest term in the set $\{|Y(x)-Y(x_{0})|:x\in \Omega
_{x_{0},d}^{0}\}$, and $2\leq K<\mathrm{card\ \Omega _{x_{0},d}^{0}}$. Let
\begin{equation}
J(x,H)=\exp \left( -\frac{ROADG(x)^{2}}{H^{2}}\right) ,
\label{impulse J function}
\end{equation}
be a weight to estimate whether the point is impulse one,
where the parameter $H$ controls the shape of the function. If the pixel $x$
is an impulse point then $ROADG(x)$ is large and $J(x,H)$ $\approx 0;$
otherwise $ROADG(x)\approx 0$ and $J(x,H)\approx 1.$

Now, we modify the Optimal Weights Filter \citep{jin2011removing} in
order to treat the mixture of impulse and Gaussian noises. Similar to (\ref%
{distance l2}), we define the impulse detection distance  by%
\begin{equation*}
d_{J,\kappa}\left( \mathbf{Y}_{x,\eta },\mathbf{Y}_{x_{0},\eta }\right) =\frac{%
\left\Vert \left( \mathbf{Y}_{x,\eta }-\mathbf{Y}_{x_{0},\eta }\right)
\right\Vert _{J,\kappa}}{\sqrt{\sum_{y^{\prime }\in \mathbf{V}_{x_{0},\eta
}}\kappa(y^{\prime })}},
\end{equation*}%
where
\begin{equation*}
\begin{split}
&\Vert \mathbf{Y}_{x,\eta }-\mathbf{Y}_{x_{0},\eta }\Vert
_{J,\kappa}^{2}
\\&
=\sum_{y\in \mathbf{V}_{x_{0},\eta
}}\kappa(T_x y)J(T_x y,H_{1})J(  y  ,H_{1})(Y(T_x y)-Y(  y  ))^{2},
\end{split}
\end{equation*}%
and $\kappa$ are some weights defined on $\mathbf{V}_{x_{0},\eta }.$ The
corresponding estimate of brightness variation $\rho _{f,x_{0}}\left(
x\right) $ is given by%
\begin{equation}
\widehat{\rho }_{J,\kappa,x_{0}}(x)=\left( {d_{J,\kappa}\left( \mathbf{Y}_{x,\eta },%
\mathbf{Y}_{x_{0},\eta }\right) }-\sqrt{2}\sigma \right) ^{+}.
\label{empir simil func}
\end{equation}%
The smoothing kernels $\kappa$ used in the simulations
are the Gaussian kernel
\begin{equation}
\kappa_{g}(y)=\exp \left( -\frac{N^{2}\Vert y-x_{0}\Vert _{2}^{2}}{2h_{g}}\right)
,  \label{s4kg}
\end{equation}%
where $h_{g}$ is the bandwidth parameter and
the following kernel: for $y \in \mathbf{U}_{x_0,\eta}$,
\begin{equation}
\kappa_{0}\left( y\right) =\sum_{k=\max(1,j)}^{\eta}\frac{1}{(2k+1)^2}
\label{s4ky}
\end{equation}%
if $\|y-x_0\|_{\infty}=j$ for some $j\in \{0,1,\cdots,\eta\}$.
 In the simulations
presented below we use the kernel $\kappa=\kappa_{0}.$

Now,  we define a new filter, called \textit{Optimal Weights Mixed Filter%
} (OWMF), by
\begin{equation}
\widehat{f}_{h }(x_{0})=\frac{\sum_{x\in \mathbf{U%
}_{x_{0},h}}J(x,H_{2})\kappa_{\text{tr}}(\frac{\widehat{\rho }_{J,\kappa,x_{0}}\left(
x\right) }{\widehat{a}_{J}})Y(x)}{\sum\limits_{y\in \mathbf{U}%
_{x_{0},h}}J(x,H_{2})\kappa_{\text{tr}}(\frac{\widehat{\rho }_{J,\kappa,x_{0}}\left(
x\right) }{\widehat{a}_{J}})},  \label{OWMF}
\end{equation}%
where the bandwidth $\widehat{a}_{J}>0$ can be calculated as in Remark \ref%
{calculate a} (with $\widehat{\rho }_{x_{0}}(x)$ and $\widehat{a}$ replaced
by $\widehat{\rho }_{J,\kappa,x_{0}}(x)$ and $\widehat{a}_{J}$ respectively) and $%
H_{2}$ is a parameter. Notice that $H_{1}$ and $H_{2}$ (which is used in the
definition of $\widehat{\rho }_{J,\kappa,x_{0}}\left( x\right) $) may take
different values.

To explain the new algorithm (\ref{OWMF}), note that the function $%
J(x,H_{2}) $ acts as a filter of the points contaminated by the impulse
noise. In fact, if $x$ is an impulse noisy point, then $J(x,H_{2})%
\thickapprox 0.$ When the impulse noisy points are filtered, the remaining
part of the image is treated as a image distorted by solely the Gaussian
noise. So, in the new filter, the basic idea is to apply the OWF %
\citep{jin2011removing} by giving nearly $0$ weights to impulse noisy points.

\section{Simulation and comparisons \label{sec Simulation and comparisons}}

The performance of a filter $\widehat{f}$ is measured by the usual Peak
Signal-to-Noise Ratio (PSNR) in decibels (db) defined by%
$$
PSNR=10\log _{10}\frac{255^{2}}{MSE},
$$
$$
MSE=\frac{1}{\mathrm{card}\,\mathbf{I}}%
\sum\limits_{x\in \mathbf{I}}(f(x)-\widehat{f}_h(x))^{2},
$$
where $f$ is the original image.

In the simulations, to avoid the undesirable border effects in our simulations, we mirror the
image outside the image limits. In more detail, we extend the image outside the
image limits symmetrically with respect to the border. At the corners, the
image is extended symmetrically with respect to the corner pixels.

\bigskip

\textbf{Algorithm :}\quad Optimal Weights Mixed Filter

For each $x\in \mathbf{I}$

\quad \quad compute $ROADG(x)=\left(\frac{1}{K}\sum_{i=1}^{K}
r_{i}(x)-\sigma\right)^+$

\quad \quad compute $J(x,H_1)=\exp\left(-\frac{ROADG(x)^2}{H_1^2}\right)$

\quad \quad compute $J(x,H_2)=\exp\left(-\frac{ROADG(x)^2}{H_2^2}\right)$

Repeat for each $x_0\in \mathbf{I}$

\quad \quad give an initial value of $\widehat {a}$: $\widehat{a}=1$ (it can
be an arbitrary positive number). 

\quad \quad compute $\{\widehat{\rho }_{J,\kappa,x_{0}}(x) :  x\in \mathbf{U}%
_{x_0,h}\}$ by (\ref{empir simil func})

/\emph{compute the bandwidth }$\widehat{a}$\emph{\ at }$x_{0}$

\quad \quad reorder $\{\widehat{\rho }_{J,\kappa,x_{0}}(x) :  x\in \mathbf{U}%
_{x_0,h}\}$ as increasing sequence, say

\quad \quad\quad \quad $\widehat{\rho }_{J,\kappa,x_{0}}(x_1)\leq\widehat{\rho }_{J,\kappa,x_{0}}(x_2)\leq \cdots \leq \widehat{\rho }_{J,\kappa,x_{0}}(x_M)$ \ \ \ \

\quad \quad loop from $k=1$ to $M$

\quad \quad \quad \quad if $\sum_{i=1}^{k}\widehat{\rho }_{J,\kappa,x_{0}}(x_{i})>0$

\quad \quad \quad \quad \quad \quad if $\frac{\sigma ^{2}+\sum_{i=1}^{k}%
\widehat{\rho }_{J,\kappa,x_{0}}^{2}(x_{i})}{\sum_{i=1}^{k}\widehat{\rho }_{J,\kappa,x_{0}}(x_{i})}%
\geq \widehat{\rho }_{J,\kappa,x_{0}}(x_{k})$ then $\widehat{a}=\frac{\sigma
^{2}+\sum_{i=1}^{k}\widehat{\rho }_{J,\kappa,x_{0}}(x_{i})}{\sum_{i=1}^{k}\widehat{\rho }_{J,\kappa,x_{0}}(x_i)}$

\quad \quad \quad \quad \quad \quad else quit loop

\quad \quad \quad \quad else continue loop

\quad \quad end loop

/\emph{compute the estimated weights }$\widehat{w}$\emph{\ at }$x_{0}$

\quad \quad compute $\widehat{w}(x_{i})=\frac{J(x,H_2)\kappa_{\text{tr}}\left(\frac{1-%
\widehat{\rho }_{x_0}(x_{i})}{\widehat{a}}\right)^{+}}{\sum_{x_{i}\in \mathbf{U}%
_{x_{0},h}}J(x,H_2)\kappa_{\text{tr}}\left(\frac{1-\widehat{\rho }_{x_0}(x_{i})}{\widehat{a}%
}\right)^{+}}$

/\emph{compute the filter }$\widehat{f}_h$\emph{\ at }$x_{0}$

\quad \quad compute $\widehat{f}_h(x_{0})=\sum_{x_{i}\in \mathbf{U}_{x_{0},h}}%
\widehat{w}(x_{i})Y(x_{i})$.

\bigskip

\begin{table*}[tbp]
\caption{Performance of denoising algorithms when applied to test Gaussian white noisy  images.
 }
\label{Table gaussian}
\begin{center}
\renewcommand{\arraystretch}{0.6} \vskip3mm {\fontsize{8pt}{\baselineskip}%
\selectfont
\begin{tabular}{|c|l|cccc|}
\hline
            &Images & Lena & Barbara & Boat & House
             \\
            &Sizes & $512 \times 512$ & $512 \times 512$ & $512 \times 512$ & $256 \times
            256$  \\ \hline\hline
$\sigma$    &         Method  & PSNR    &  PSNR   &  PSNR   &  PSNR      \\ \hline
                        &Our method&&&&\\
                        &$M=13\times13$
            & 33.75db & 31.81db & 31.02db & 33.82db  \\
                        &$m=25\times25$&&&&\\ \cline{2-6}
                      \citet{buades2005non}
            & 32.72db & 31.67db & 30.39db & 33.82db  \\
                        & \citet{katkovnik2004directional}
            & 32.18db & 29.20db & 30.46db & 32.62db  \\
$15$                    &\citet{foinovel}
            & 32.72db & 29.61db & 30.93db & 33.18db  \\
                        & \citet{roth2009fields}
            & 33.29db & 30.16db & 31.27db & 33.55db  \\
                        & \citet{hirakawa2006image}
            & 33.97db & 32.55db & 31.59db & 33.82db  \\
                        &\citet{kervrann2008local}
            & 33.70db & 31.80db & 31.44db & 34.08db  \\
                        & \citet{jin2011removing}
            & 33.93db & 32.31db & 31.64db & 34.09db  \\
                        & \citet{hammond2008image}
            & 34.04db & 32.25db & 31.72db & 33.72db  \\
                        & \citet{aharon2006rm}
            & 33.71db & 32.41db & 31.77db & 34.25db  \\
                        &\citet{dabov2007image}
            & 34.27db & 33.00db & 32.14db & 34.94db  \\
            \hline
                        &Our method&&&&\\
                        &$M=13\times13$
            & 32.42db & 30.40db & 29.62db & 32.71db  \\
                        &$m=27\times27$&&&&\\ \cline{2-6}
                        &\citet{buades2005non}
            & 31.51db & 30.38db & 29.32db & 32.51db  \\
                        & \citet{katkovnik2004directional}
            & 30.74db & 27.38db & 29.03db & 31.24db  \\
$20$                    & \citet{foinovel}
            & 31.43db & 27.90db & 39.61db & 31.84db  \\
                        & \citet{roth2009fields}
            & 31.89db & 28.28db & 29.86db & 32.29db  \\
                        & \citet{hirakawa2006image}
            & 32.69db & 31.06db & 30.25db & 32.58db  \\
                        & \citet{kervrann2008local}
            & 32.64db & 30.37db & 30.12db & 32.90db  \\
                        & \citet{jin2011removing}
            & 32.68db & 31.04db & 30.30db & 32.83db  \\
                        & \citet{hammond2008image}
            & 32.81db & 30.76db & 30.41db & 32.52db  \\
                        & \citet{aharon2006rm}
            & 32.39db & 30.84db & 30.39db & 33.10db  \\
                        & \citet{dabov2007image}
            & 33.05db & 31.78db & 30.88db & 33.77db  \\
            \hline
                        &Our method&&&&\\
                        &$M=13\times13$
            & 31.40db & 29.20db & 28.56db & 31.61db  \\
                        &$m=27\times27$&&&&\\ \cline{2-6}
                        & \citet{buades2005non}
            & 30.36db & 29.19db & 28.38db & 31.16db  \\
                        & \citet{katkovnik2004directional}
            & 29.66db & 26.05db & 27.93db & 30.12db  \\
$25$                    &\citet{foinovel}
            & 30.43db & 26.62db & 28.60db & 30.75db  \\
                        & \citet{roth2009fields}
            & 30.57db & 26.84db & 28.57db & 31.05db  \\
                        & \citet{hirakawa2006image}
            & 31.69db & 29.89db & 29.21db & 31.60db  \\
                        & \citet{kervrann2008local}
            & 31.73db & 29.24db & 29.20db & 32.22db  \\
                        & \citet{jin2011removing}
             & 31.59db & 29.92db & 29.16db & 31.95db \\
                        & \citet{hammond2008image}
            & 31.83db & 29.58db & 29.40db & 31.54db  \\
                        & \citet{aharon2006rm}
            & 31.36db & 29.58db & 29.32db & 32.07db  \\
                        & \citet{dabov2007image}
            & 32.08db & 30.72db & 29.91db & 32.86db  \\\hline
\end{tabular}
} \vskip1mm
\end{center}
\end{table*}

In our simulations, the parameters can be choose as follows:
\begin{equation*}
\begin{split}
d& =2, \\
K& =12, \\
M& =13\times 13, \\
m& =15\times 15, \\
H_{1}& =5+\frac{30}{1+20p}+(\sigma -10)^{+}(0.5-p), \\
H_{2}& =27-20p.
\end{split}%
\end{equation*}%
In \citep{garnett2005universal} it is suggested to take $d=1$ and $K=4$. For
low and moderate levels of noise $(p<25\%)$, one iteration is sufficient and
usually provides the best results; for high levels of noise $(p>25\%)$,
applying two to five iterations provides better results. In our simulations,
we found that a few spots of unremoved impulses often remain if we choose $%
d=1$ and $K=4$. This happens because impulses sometimes "clump" together,
and the $3\times 3$ detection window is too small to identify all the
impulse noise points. Consequently, we select parameters $d=2$ and $K=12$ of
detection windows for all levels of impulse noise. Figure \ref{comparison}
shows the comparison results between the restored images, with detection
window $3\times 3$ and with detection window $5\times 5$, which have been
added an impulse noise with $p=20\%$, $30\%,$ $40\%,$ and $50\%$
respectively. When $p=30\%$, $40\%$ and $50\%$, we can see clearly some
impulse spots in the restored images with detection window $3\times 3$,
while the visual quality of the restored images with detection window $%
5\times 5$ is very good, without impulse spots. In the case where $p=20\%$,
impulse spots of the restored image with detection window $3\times 3$ are
not obvious, and the PSNR value is a little better than that with detection window $5\times5$, whereas Figure \ref%
{figure comparison part} shows that the first image has two clumpy impulse
spots and the visual quality is not good enough. Consequently, we prefer
detection window $5\times 5$ for all levels impulse noise.

\begin{table*}[tbp]
\caption{Performance of denoising algorithms when applied to test impulse noisy  images.
 }
\label{Table impulse}
\begin{center}
\renewcommand{\arraystretch}{0.5} \addtolength{\tabcolsep}{-2pt} \vskip3mm {\fontsize{8pt}{\baselineskip}%
\selectfont
\begin{tabular}{|l|cc|cc|cc|cc|}
\hline
            Images & \multicolumn{2}{c|} {Baboon} &\multicolumn{2}{c|} {Bridge} &\multicolumn{2}{c|} {Lena} &\multicolumn{2}{c|} {Pentagon}
            \\ \hline
                        p\% & $20\%$ & $40\%$ & $20\%$ & $40\%$ & $20\%$ & $40\%$ & $20\%$ & $40\%$ \\ \hline\hline
             Method  & PSNR    &  PSNR   &  PSNR   &  PSNR & PSNR    &  PSNR   &  PSNR   &  PSNR    \\ \hline
                        Our method&&&&&&&&\\
                        $M=13\times13$
                                           &24.81db&\textbf{22.12}db&27.84db&24.91db&35.50db&32.19db&\textbf{30.91}db&\textbf{28.34}db\\
                        $m=25\times25$&&&&&&&&\\ \hline
 \citet{sun1994detail}        &23.67db&20.85db&26.26db&22.66db&32.93db&27.90db&29.34db&26.26db\\
 \citet{abreu1996new}           &23.81db&21.49db&26.56db&23.80db&35.71db&29.85db&30.38db&27.27db\\
 \citet{wang1999progressive} &23.43db&21.07db&26.33db&22.75db&35.09db&28.92db&29.18db&26.19db\\
\citet{chen1999tri}             &23.73db&21.38db&26.52db&22.89db&34.21db&28.30db&29.29db&26.29db\\
 \citet{chen2001space}         &24.02db&21.52db&27.27db&23.55db&35.44db&29.26db&30.34db&27.04db\\
 \citet{chen2001adaptive}      &24.17db&21.58db&27.08db&23.23db&36.07db&28.79db&30.23db&26.84db\\
\citet{crnojevic2004advanced}
                                           &23.78db&21.56db&26.90db&23.83db&36.50db&31.41db&30.11db&27.33db\\
 \citet{wenbin2005new}                  &24.18db&21.41db&27.05db&23.88db&36.90db&30.25db&30.42db&26.93db\\
 \citet{garnett2005universal} &24.18db&21.60db&27.60db&24.01db&36.70db&31.12db&30.33db&27.14db\\
 \citet{chan2004iterative}       &23.97db&21.62db&27.31db&24.60db&36.57db&32.21db&30.03db&27.35db\\
 \citet{dong2007detection}
                                           &24.49db&21.92db&27.86db&24.79db&\textbf{37.45}db&\textbf{32.76}db&30.73db&27.73db\\
 \citet{yu2008efficient}           &\textbf{24.86}db&22.06db&\textbf{28.06}db&\textbf{24.97}db&36.18db&32.03db&-    &-
                          \\\hline
\end{tabular}
} \vskip1mm
\end{center}
\end{table*}

\begin{table*}[tbp]
\caption{Comparison between the   TriF  \citep{garnett2005universal}, and  OWMF (Our method) in removing mixed noise. }
\label{Table mixed}
\begin{center}
\renewcommand{\arraystretch}{0.5} \addtolength{\tabcolsep}{-3pt} \vskip3mm {\fontsize{8pt}{\baselineskip}%
\selectfont
\begin{tabular}{|c|c|c|cccc|}
\hline
Gaussian Noise & Image & Method & $p=0.2$ & $p=0.3$ & $p=0.4$ & $p=0.5$ \\
\hline\hline
&  &  \citet{garnett2005universal} & 31.48db & 29.87db & 28.57db & 27.31db \\
& \raisebox{1 ex}[0pt]{Lena} & Our method & 33.18db & 32.05db & 30.90db & 29.52db
\\ \cline{2-7}
&  &  \citet{garnett2005universal} & 25.82db & 24.92db & 23.79db & 22.28db \\
& \raisebox{1 ex}[0pt]{Bridge} & Our method & 26.42db & 25.19db & 24.08db & 23.08db
\\ \cline{2-7}
\raisebox{1 ex}[0pt]{sigma=10} &  &  \citet{garnett2005universal} & 28.61db & 27.54db & 26.22db &
24.74db \\
& \raisebox{1 ex}[0pt]{Boat} & Our method & 29.57db & 28.22db & 27.05db & 25.92db
\\ \cline{2-7}
&  &  \citet{garnett2005universal} & 24.82db & 24.00db & 23.08db & 22.33db \\
& \raisebox{1 ex}[0pt]{Barbara} & Our method & 28.47db & 26.46db & 24.83db &
23.62db \\ \hline\hline
&  & \citet{garnett2005universal} & 28.85db & 28.02db & 27.10db & 25.68db \\
& \raisebox{1 ex}[0pt]{Lena} & Our method & 30.87db & 30.09db & 29.19db & 28.14db
\\ \cline{2-7}
&  &  \citet{garnett2005universal} & 23.56db & 23.01db & 22.47db & 21.72db \\
& \raisebox{1 ex}[0pt]{Bridge} & Our method & 24.70db & 23.97db & 23.21db & 22.45db
\\ \cline{2-7}
\raisebox{1 ex}[0pt]{sigma=20} &  &  \citet{garnett2005universal} & 26.18db & 25.46db & 24.75db &
23.79db \\
& \raisebox{1 ex}[0pt]{Boat} & Our method & 27.79db & 26.93db & 25.97db & 25.08db
\\ \cline{2-7}
&  &  \citet{garnett2005universal} & 23.35db & 22.95db & 22.53db & 21.84db \\
& \raisebox{1 ex}[0pt]{Barbara} & Our method & 27.50db & 25.95db & 24.43db &
23.33db \\ \hline\hline
&  &  \citet{garnett2005universal} & 27.26db & 26.57db & 25.58db & 23.99db \\
& \raisebox{1 ex}[0pt]{Lena} & Our method & 29.12db & 28.49db & 27.76db & 26.75db
\\ \cline{2-7}
&  &  \citet{garnett2005universal} & 22.88db & 22.42db & 21.87db & 20.98db \\
& \raisebox{1 ex}[0pt]{Bridge} & Our method & 23.56db & 23.02db & 22.49db & 21.86db
\\ \cline{2-7}
\raisebox{1 ex}[0pt]{sigma=30} &  &  \citet{garnett2005universal} & 25.11db & 24.55db & 23.80db &
22.62db \\
& \raisebox{1 ex}[0pt]{Boat} & Our method & 26.41db & 25.79db & 25.08db & 24.26db
\\ \cline{2-7}
&  &  \citet{garnett2005universal} & 22.82db & 22.46db & 21.94db & 21.10db \\
& \raisebox{1 ex}[0pt]{Barbara} & Our method & 25.98db & 24.81db & 23.72db &
22.81db \\ \hline
\end{tabular}
} \vskip1mm
\end{center}
\end{table*}

\begin{figure}[tbp]
\begin{center}
\renewcommand{\arraystretch}{0.5} \addtolength{\tabcolsep}{-6pt} \vskip3mm {%
\fontsize{8pt}{\baselineskip}\selectfont
\begin{tabular}{cccc}
$p=20\%$ & $p=30\%$ & $p=40\%$ & $p=50\%$ \\
\includegraphics[width=0.23\linewidth]{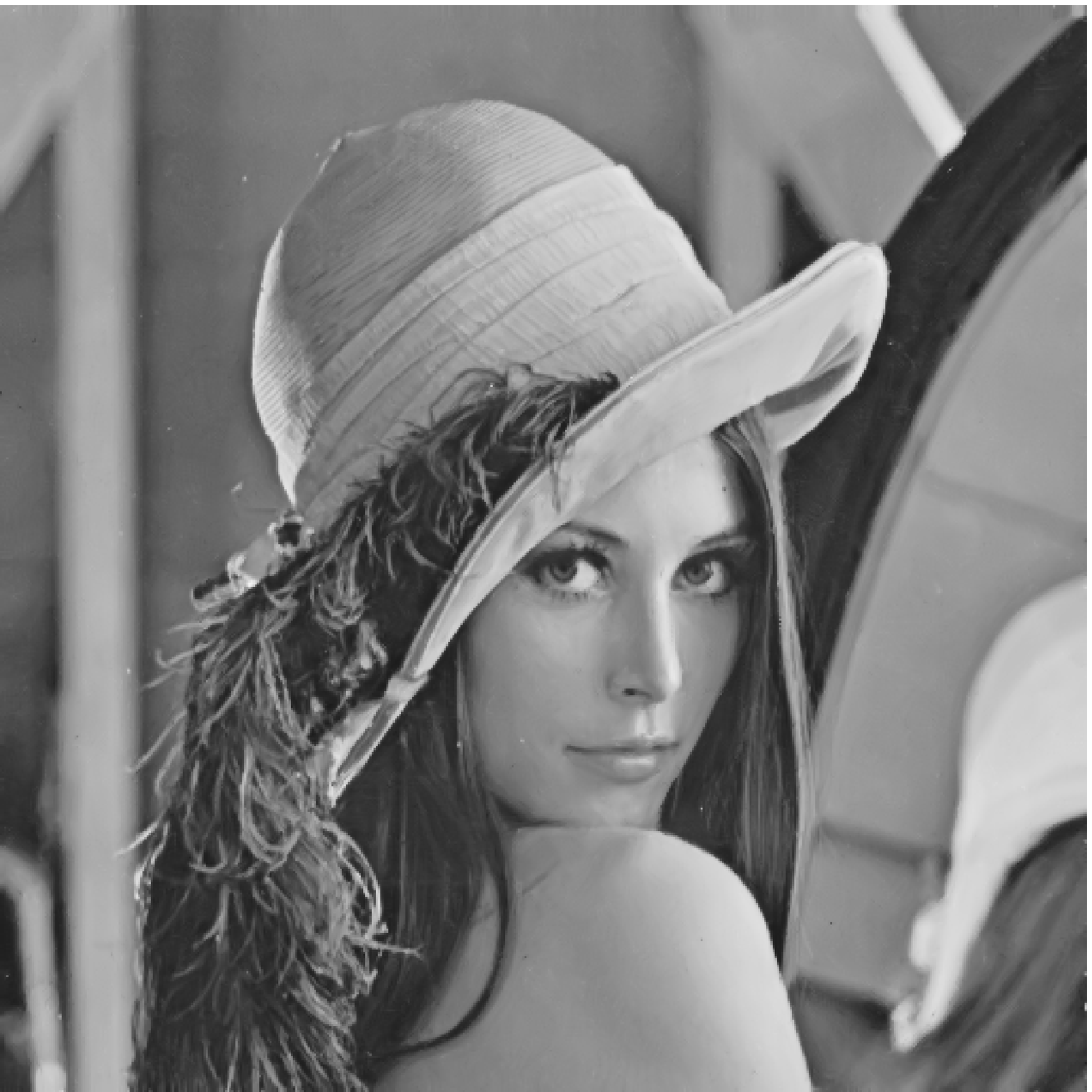} & %
\includegraphics[width=0.23\linewidth]{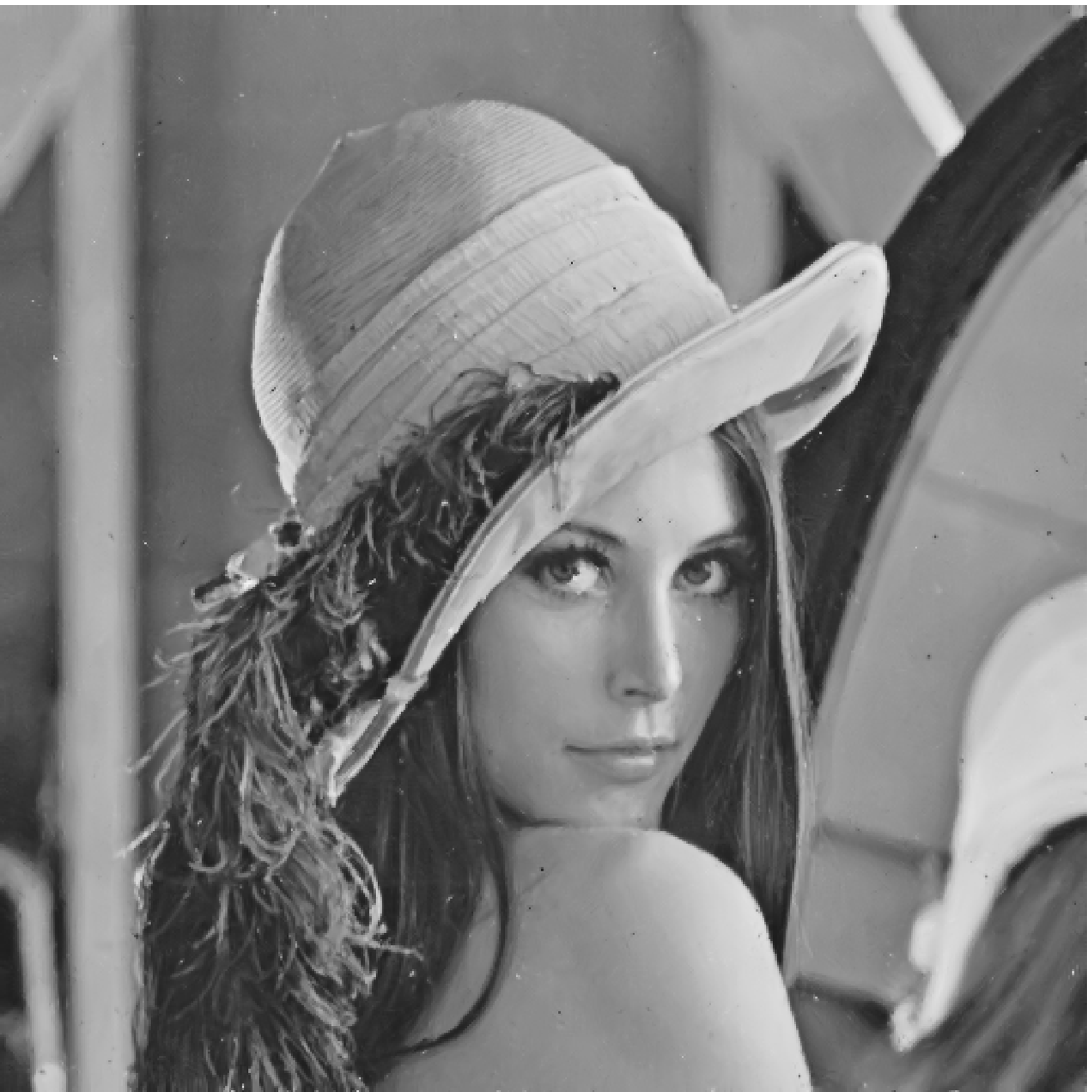} & %
\includegraphics[width=0.23\linewidth]{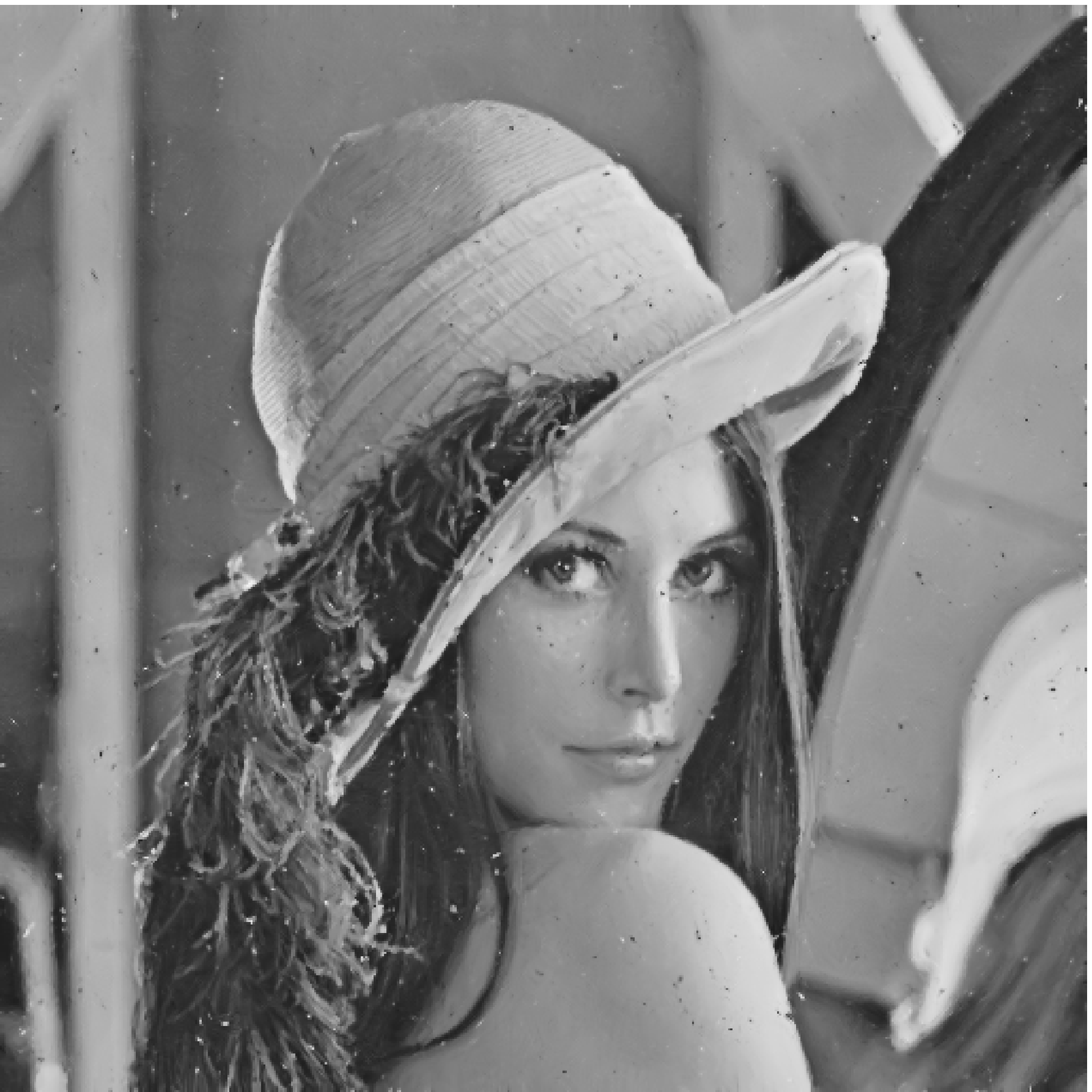} & %
\includegraphics[width=0.23\linewidth]{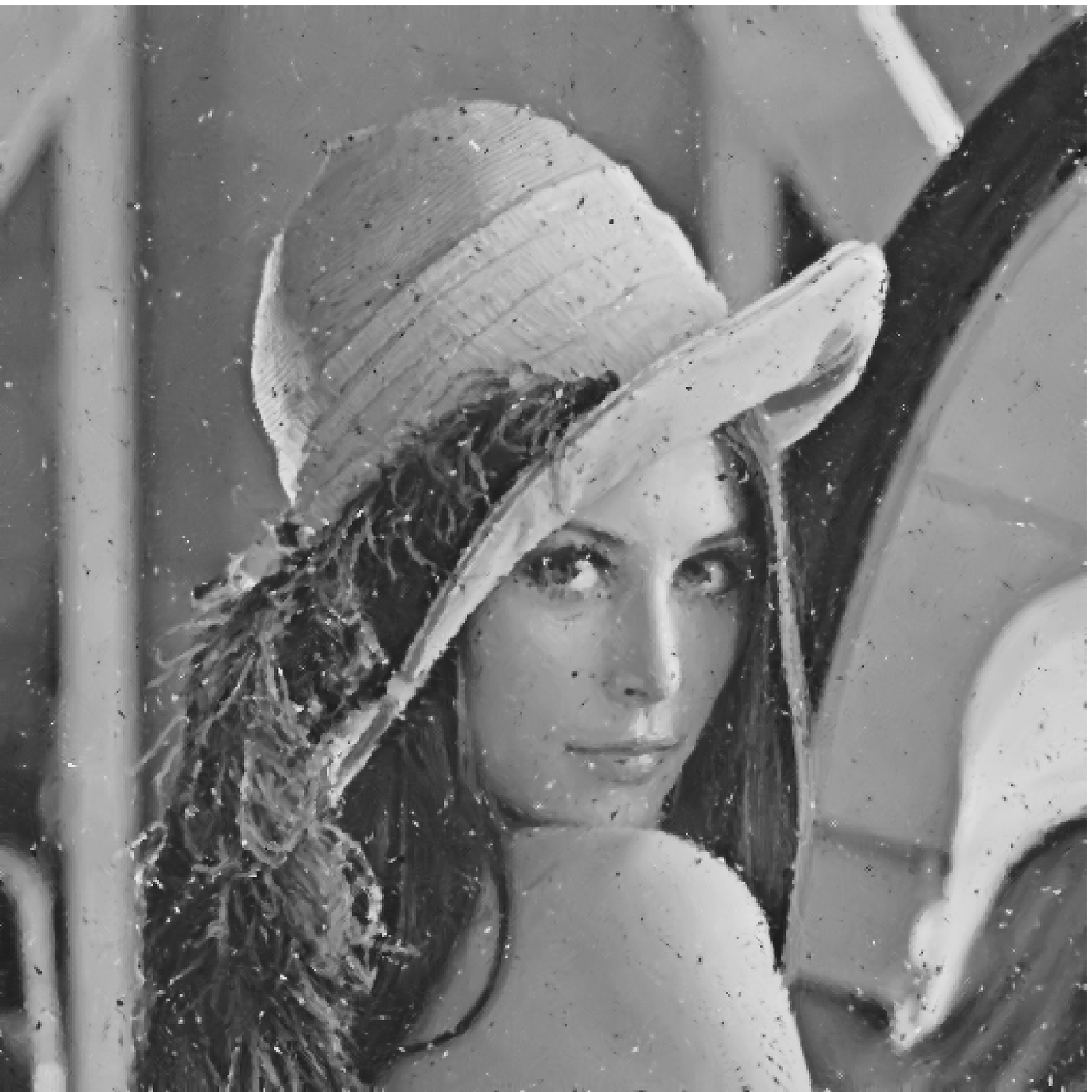} \\
PSNR=36.03db & PSNR=33.65db & PSNR=30.22db & PSNR=27.35db \\
\includegraphics[width=0.23\linewidth]{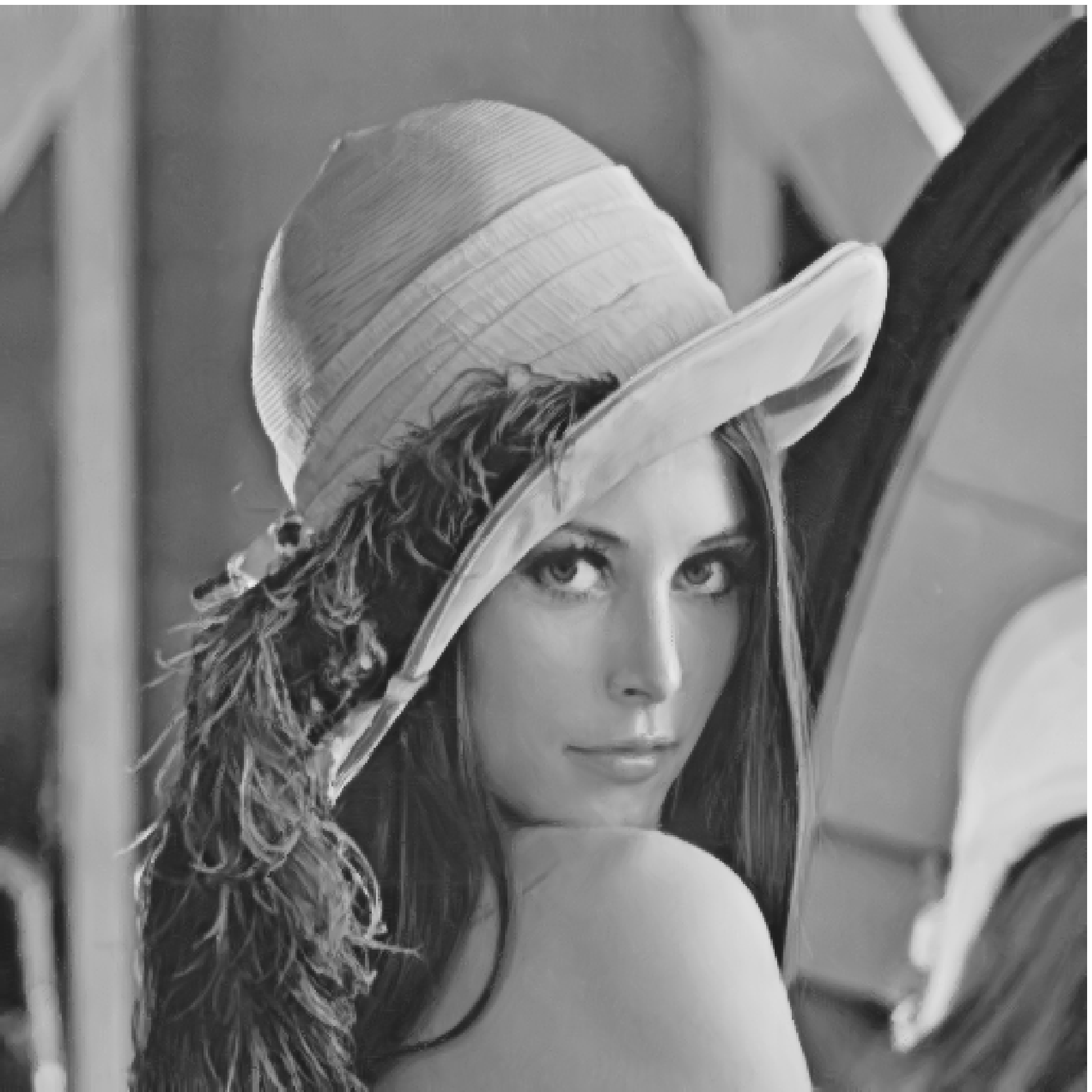} & %
\includegraphics[width=0.23\linewidth]{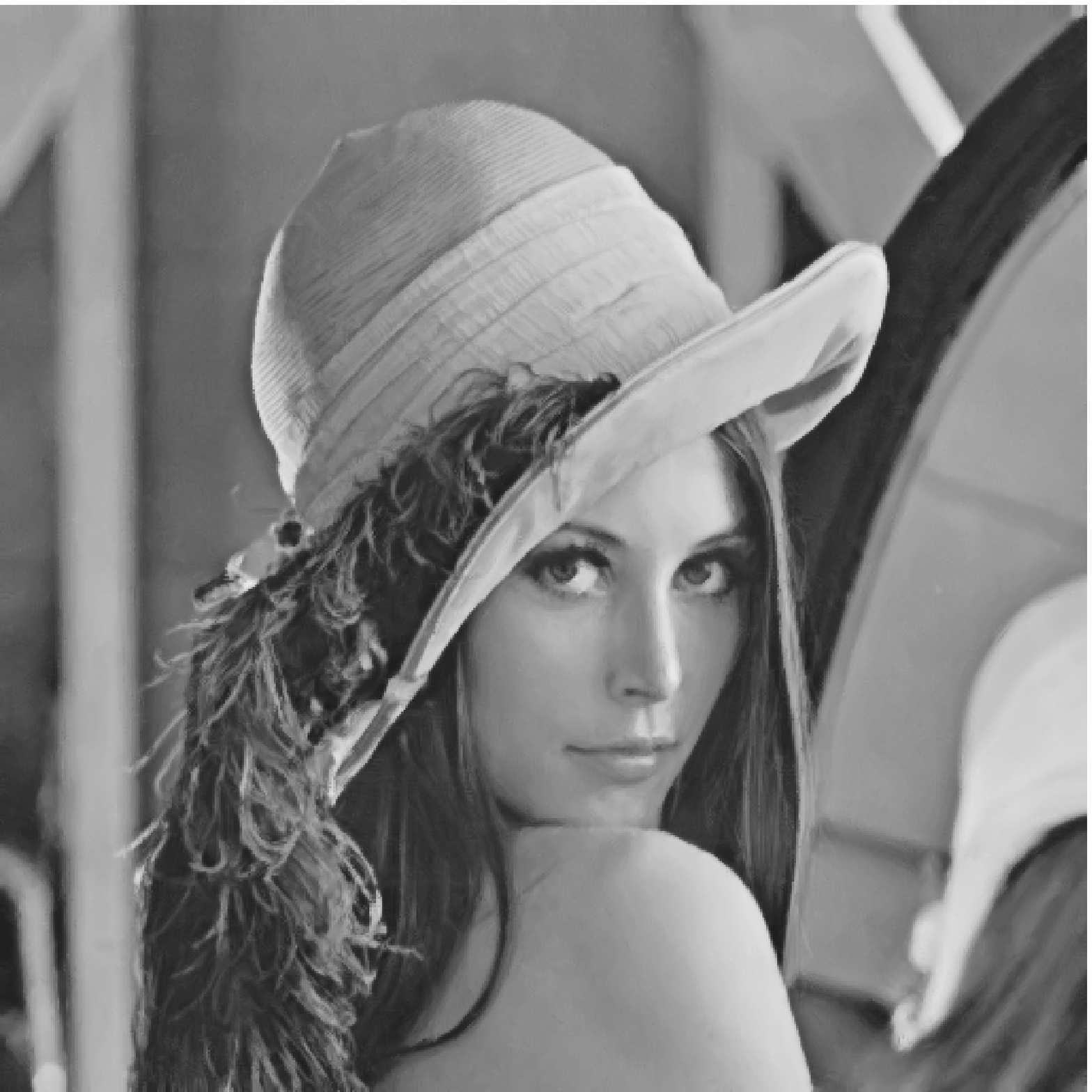} & %
\includegraphics[width=0.23\linewidth]{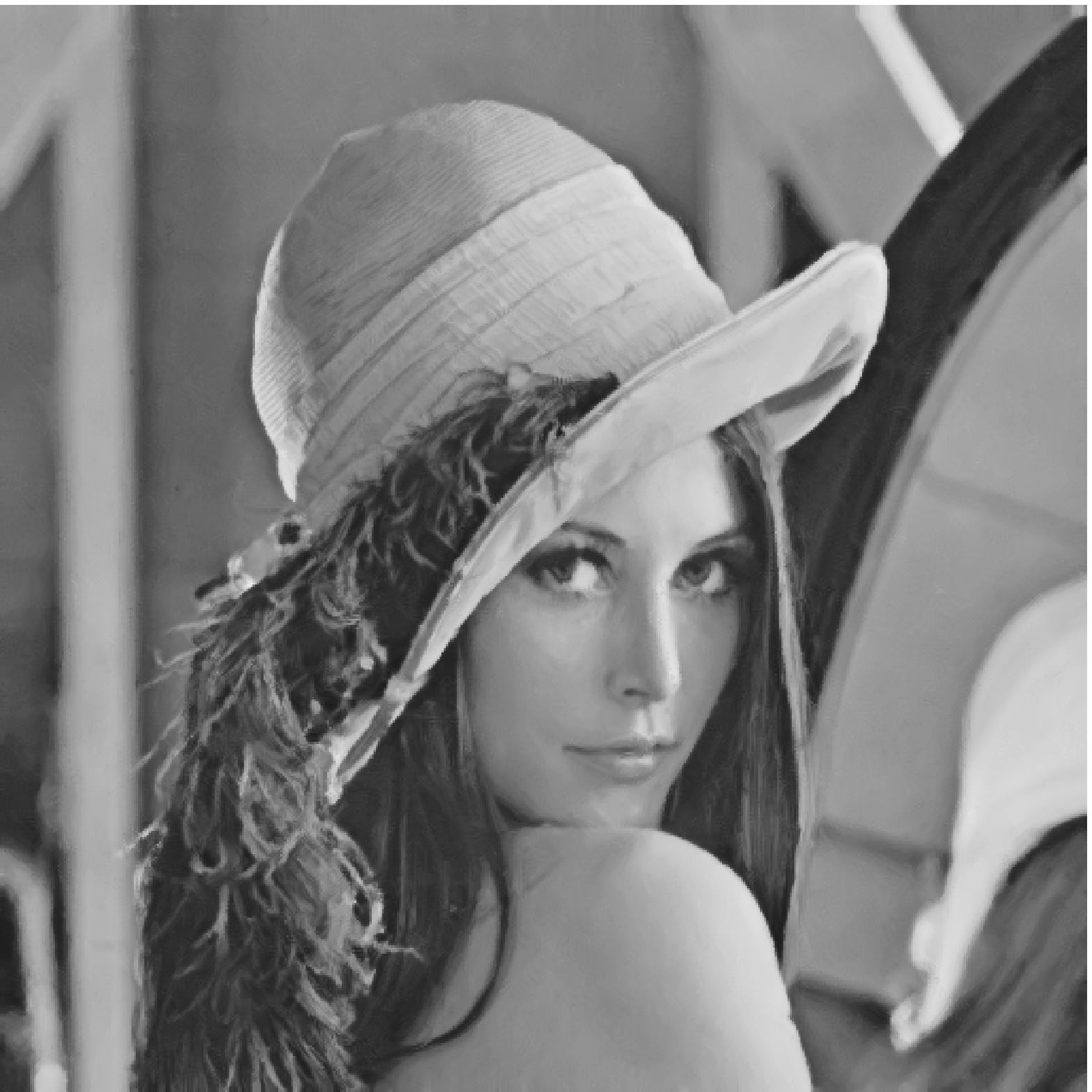} & %
\includegraphics[width=0.23\linewidth]{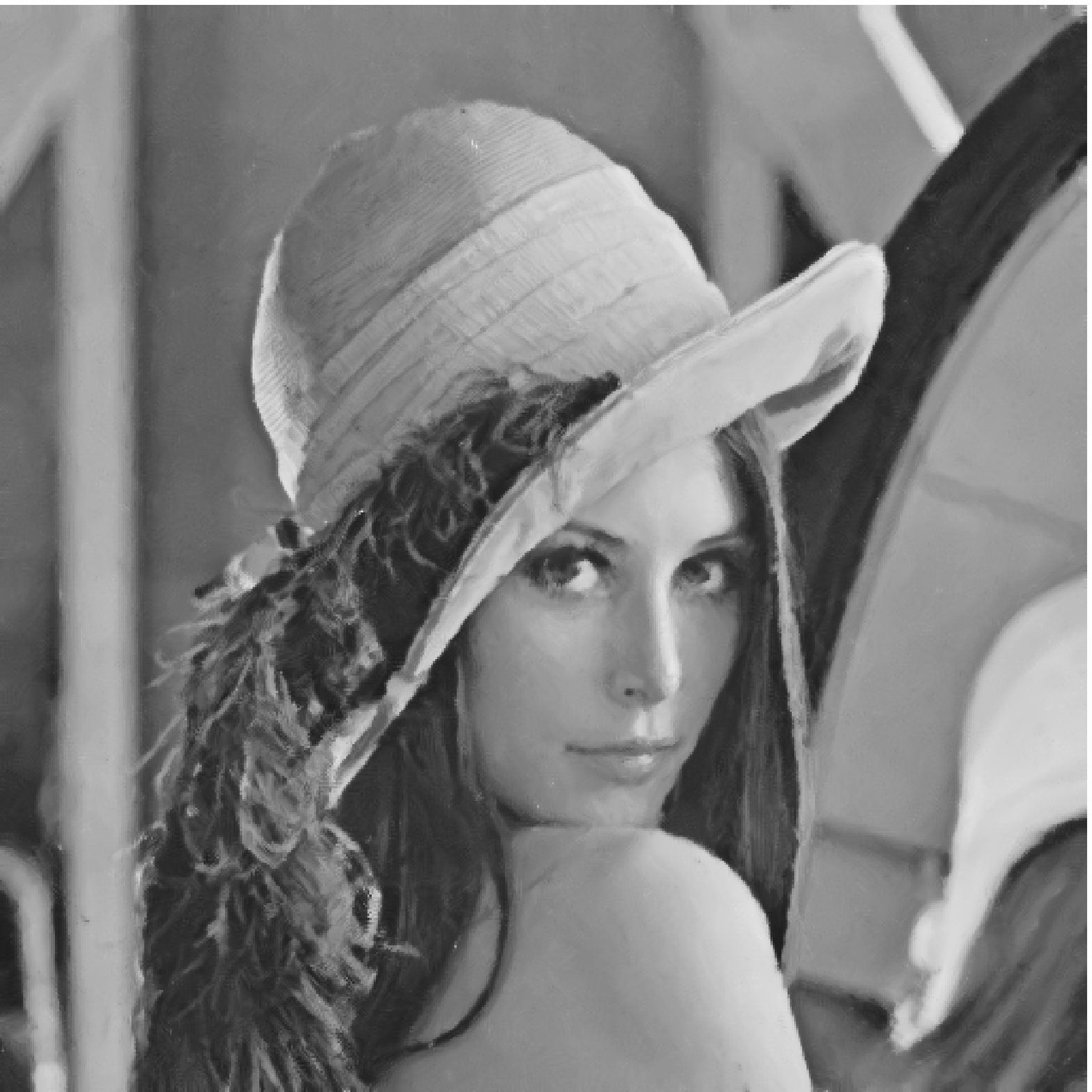} \\
PSNR=35.50db & PSNR=33.92db & PSNR=32.19db & PSNR=30.09db%
\end{tabular}
} \vskip1mm
\par
\rule{0pt}{-0.2pt}%
\par
\vskip1mm 
\end{center}
\caption{{\protect\small The first row gives the levels of added impulse
noise. The second and the third rows display images restored with detection
window $3\times3$, and their PSNR values. The forth and fifth rows show
images restored with detection window $5\times5$ and their PSNR values . The 1,
2, 3 and 4 columns give the restored images which have been contaminated by an impulse
noise with $p=20\%$, $30\%,$ $40\%,$ and $50\%$ respectively.}}
\label{comparison}
\end{figure}

\begin{figure}[tbp]
\begin{center}
\renewcommand{\arraystretch}{0.5} \addtolength{\tabcolsep}{-5pt} \vskip3mm {%
\fontsize{8pt}{\baselineskip}\selectfont
\begin{tabular}{cccc}
\includegraphics[width=0.40\linewidth]{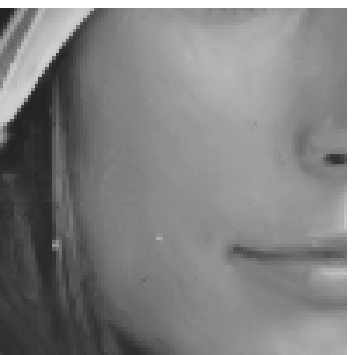} & %
\includegraphics[width=0.40\linewidth]{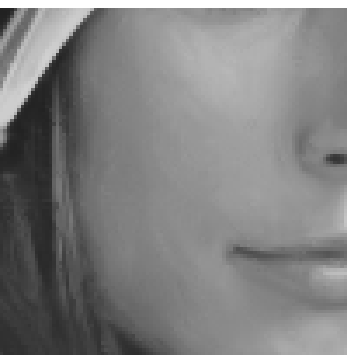} &  &
\end{tabular}
} \vskip1mm
\par
\rule{0pt}{-0.2pt}%
\par
\vskip1mm 
\end{center}
\caption{{\protect\small The first and second $100\times100$ images are
parts of images restored with detection windows $3\times3$ and
$5\times5$ respectively; the original image has been contaminated by an impulse noise with $%
p=20\%$.}}
\label{figure comparison part}
\end{figure}

\begin{figure}[tbp]
\begin{center}
\renewcommand{\arraystretch}{0.2} \addtolength{\tabcolsep}{-5pt} \vskip3mm {%
\fontsize{8pt}{\baselineskip}\selectfont
\begin{tabular}{ccccc}
$p=20$, $\sigma=0$ & $p=20$, $\sigma=10$ & $p=0$, $\sigma=20$ &    \\
\includegraphics[width=0.30\linewidth]{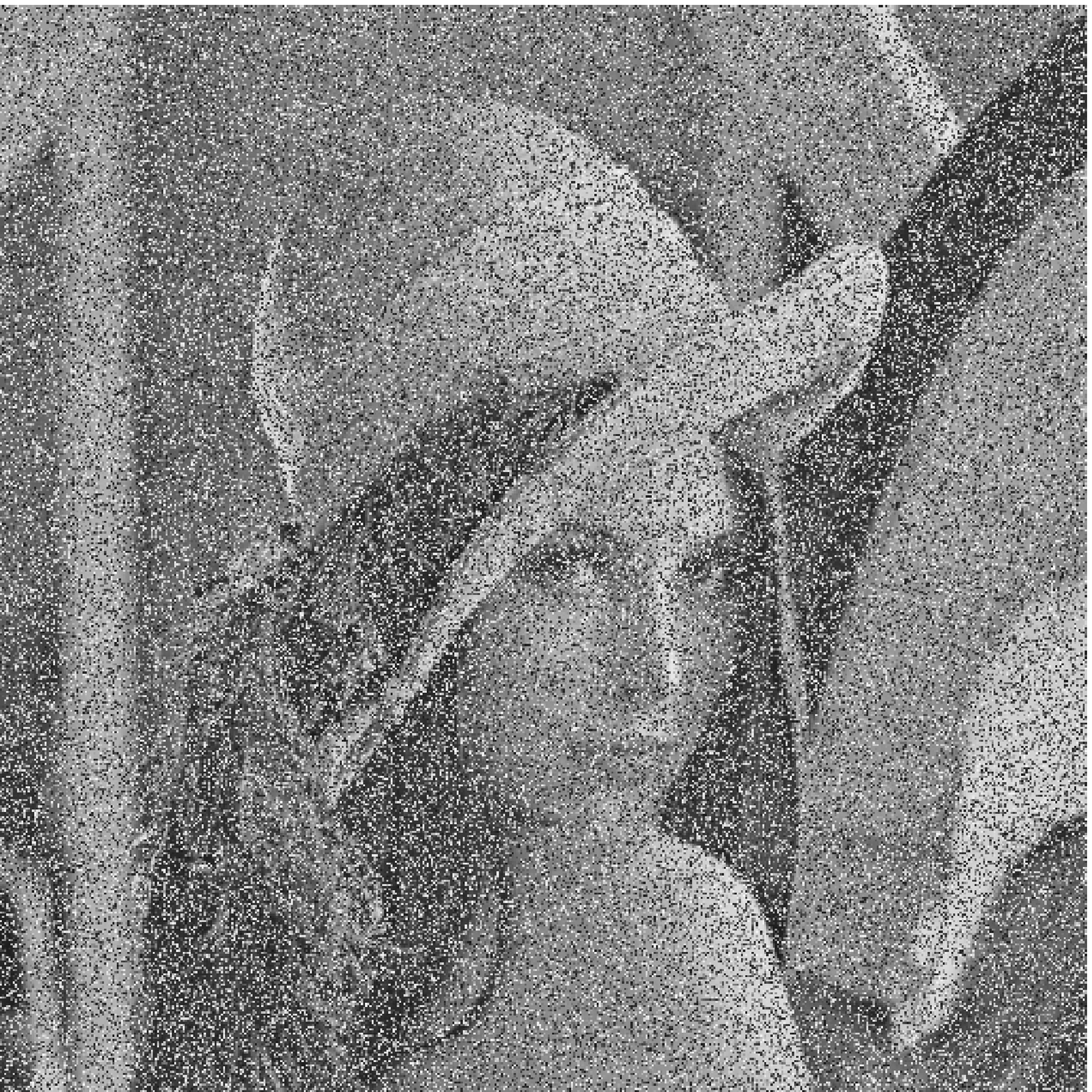} & %
\includegraphics[width=0.30\linewidth]{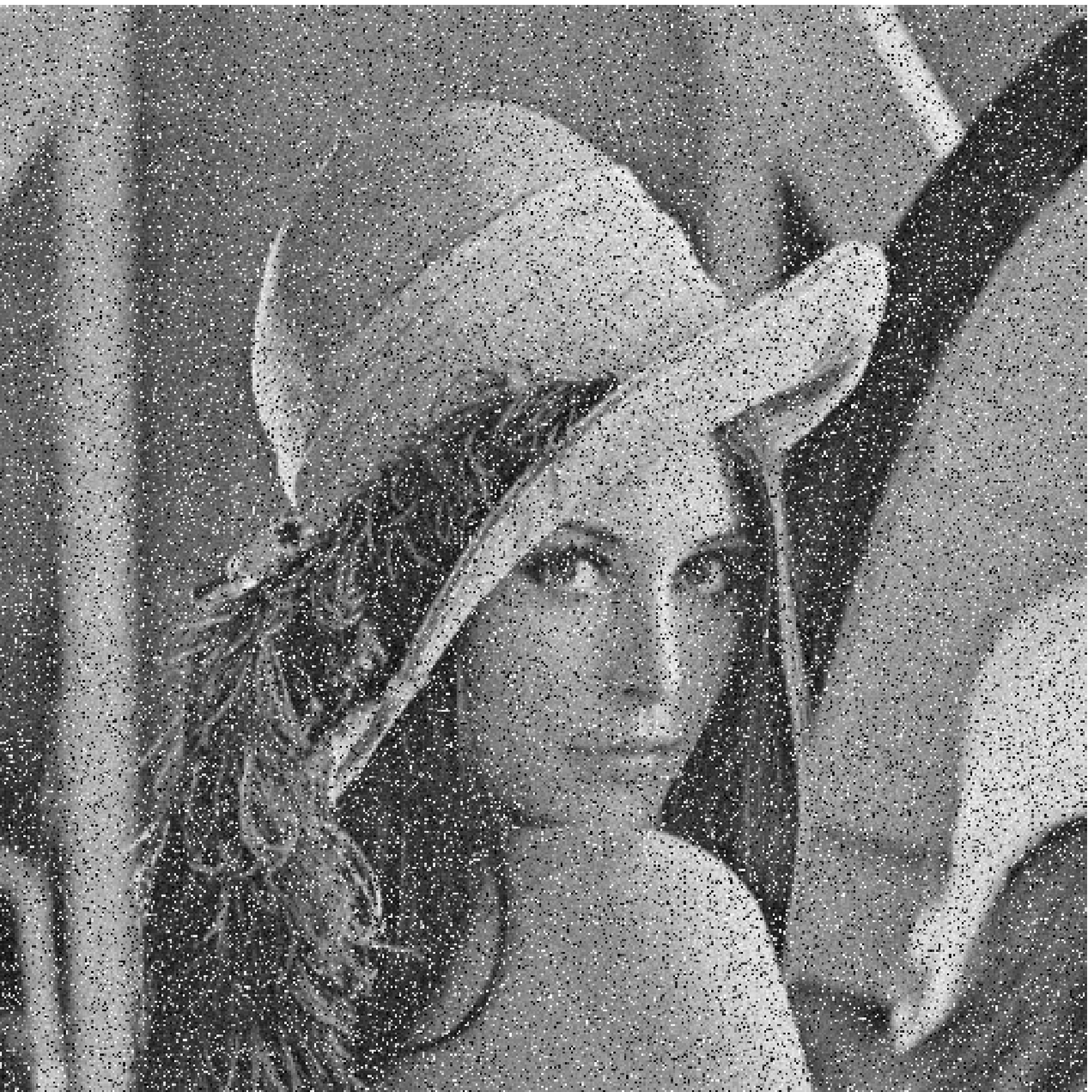} & %
\includegraphics[width=0.30\linewidth]{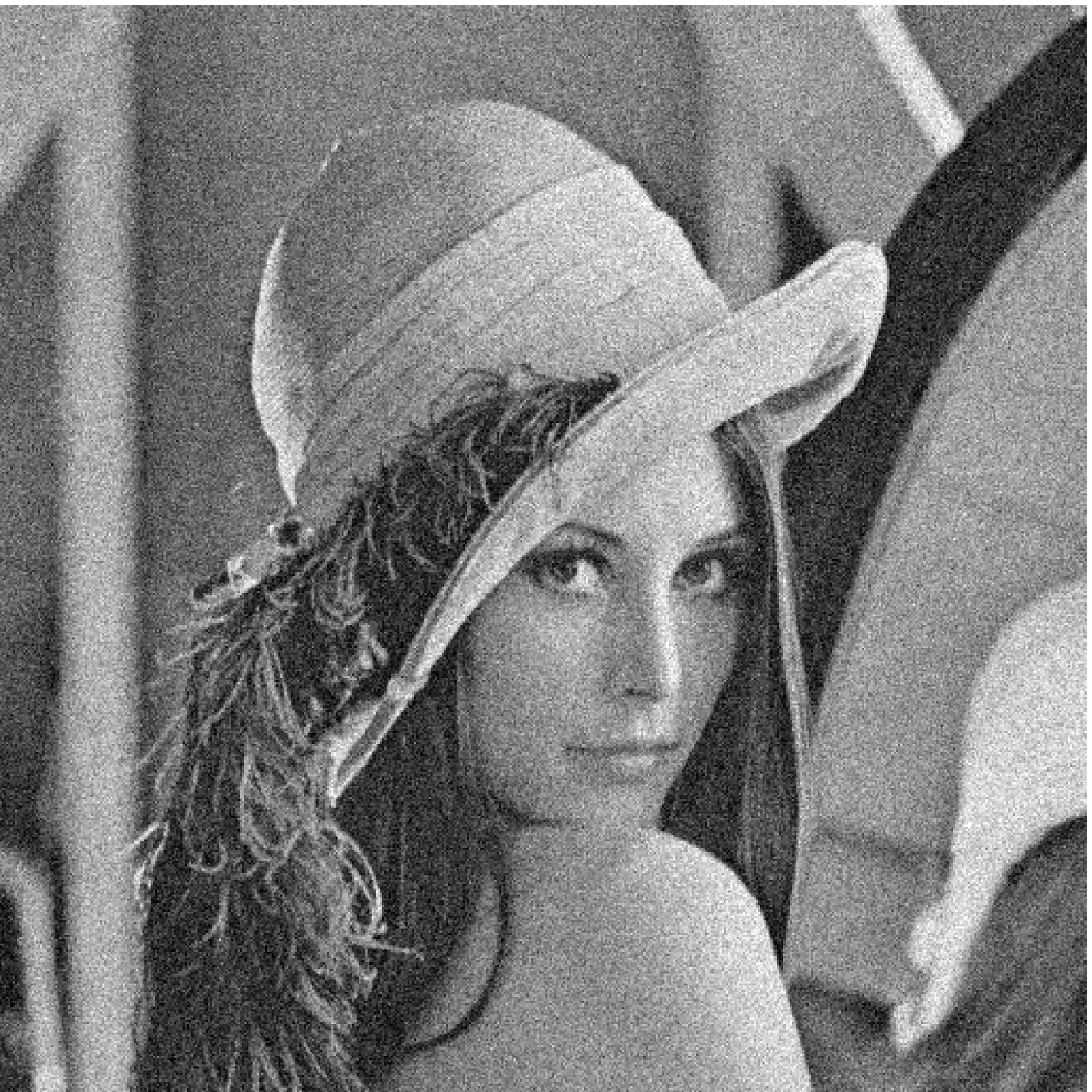} &    \\
\includegraphics[width=0.30\linewidth]{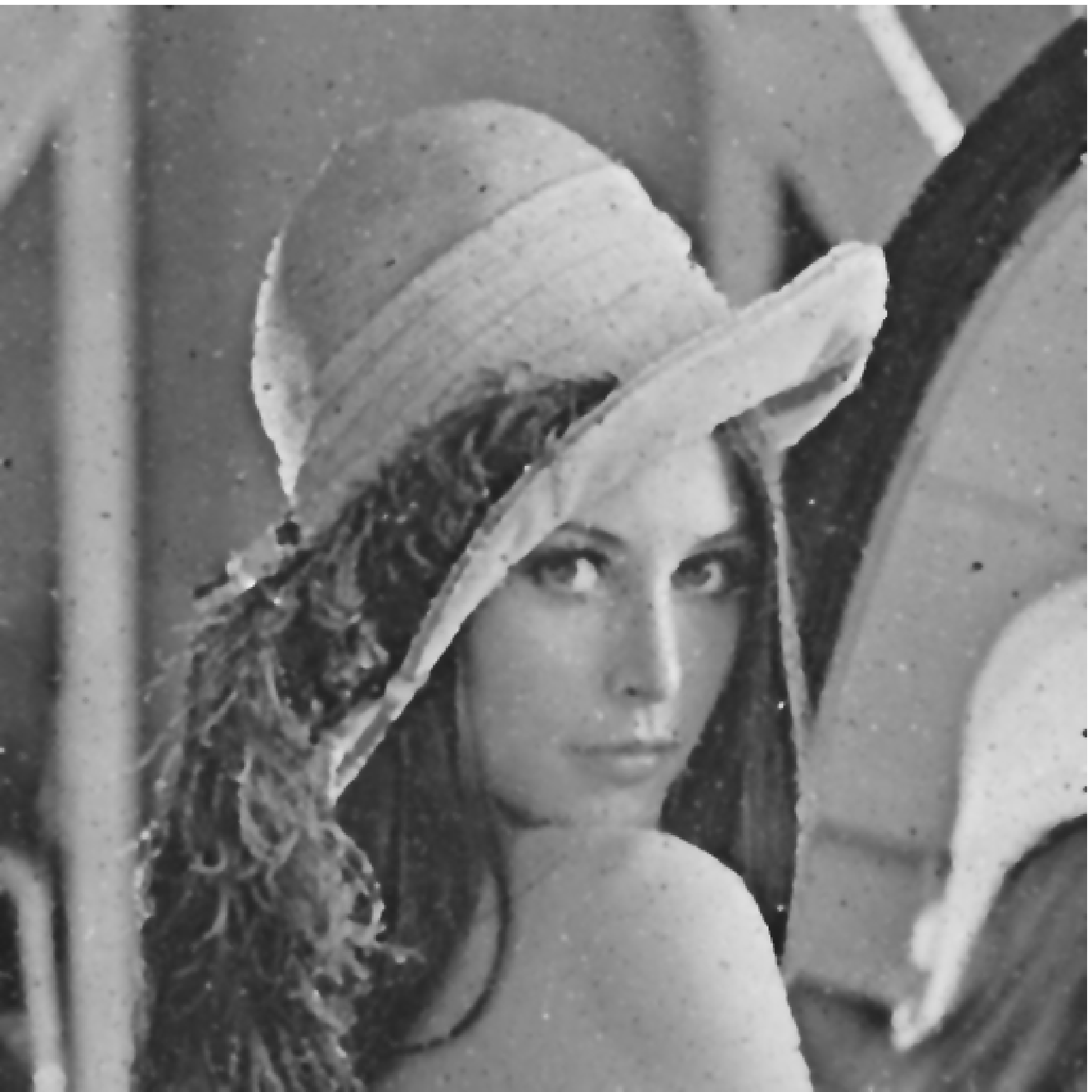} & %
\includegraphics[width=0.30\linewidth]{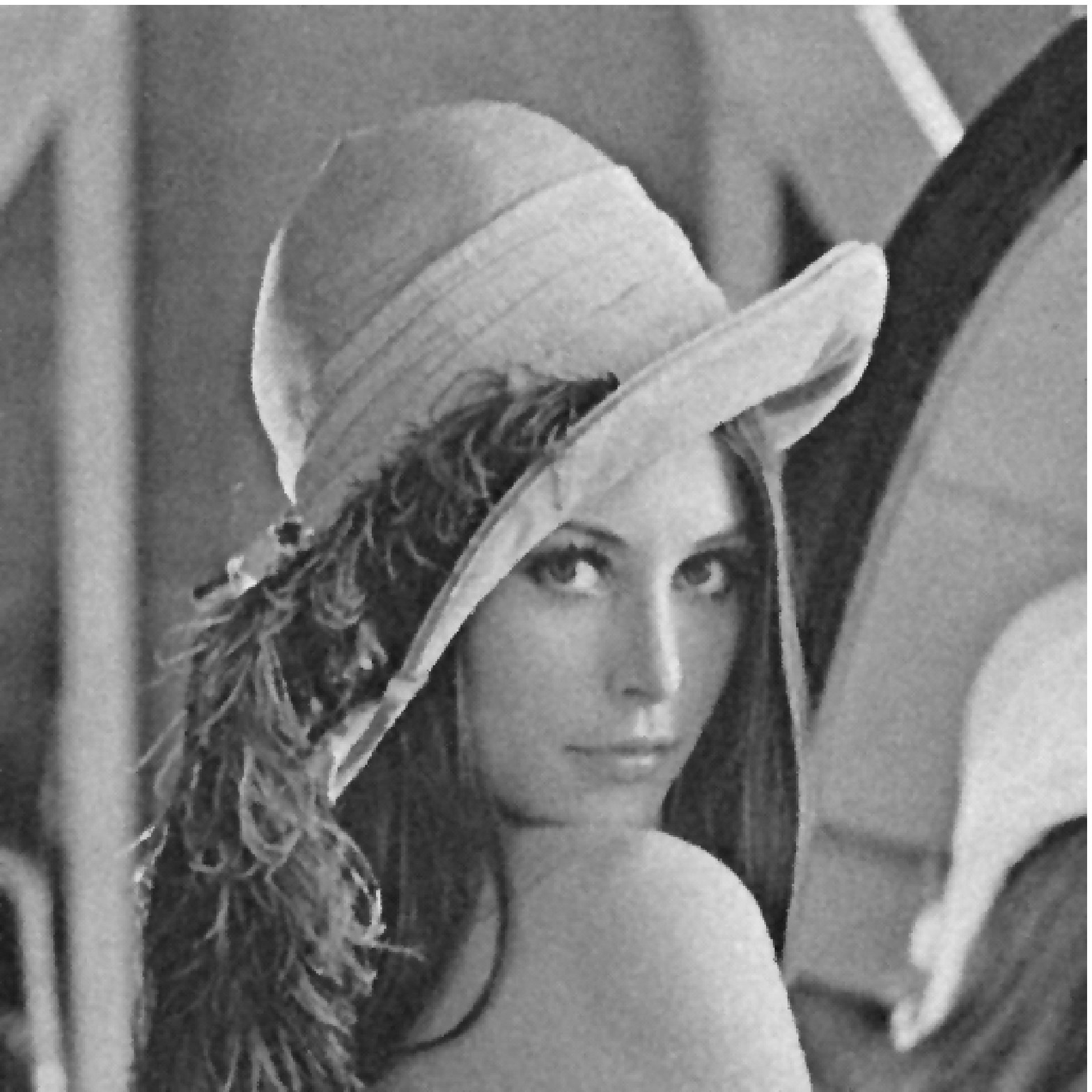} & %
\includegraphics[width=0.30\linewidth]{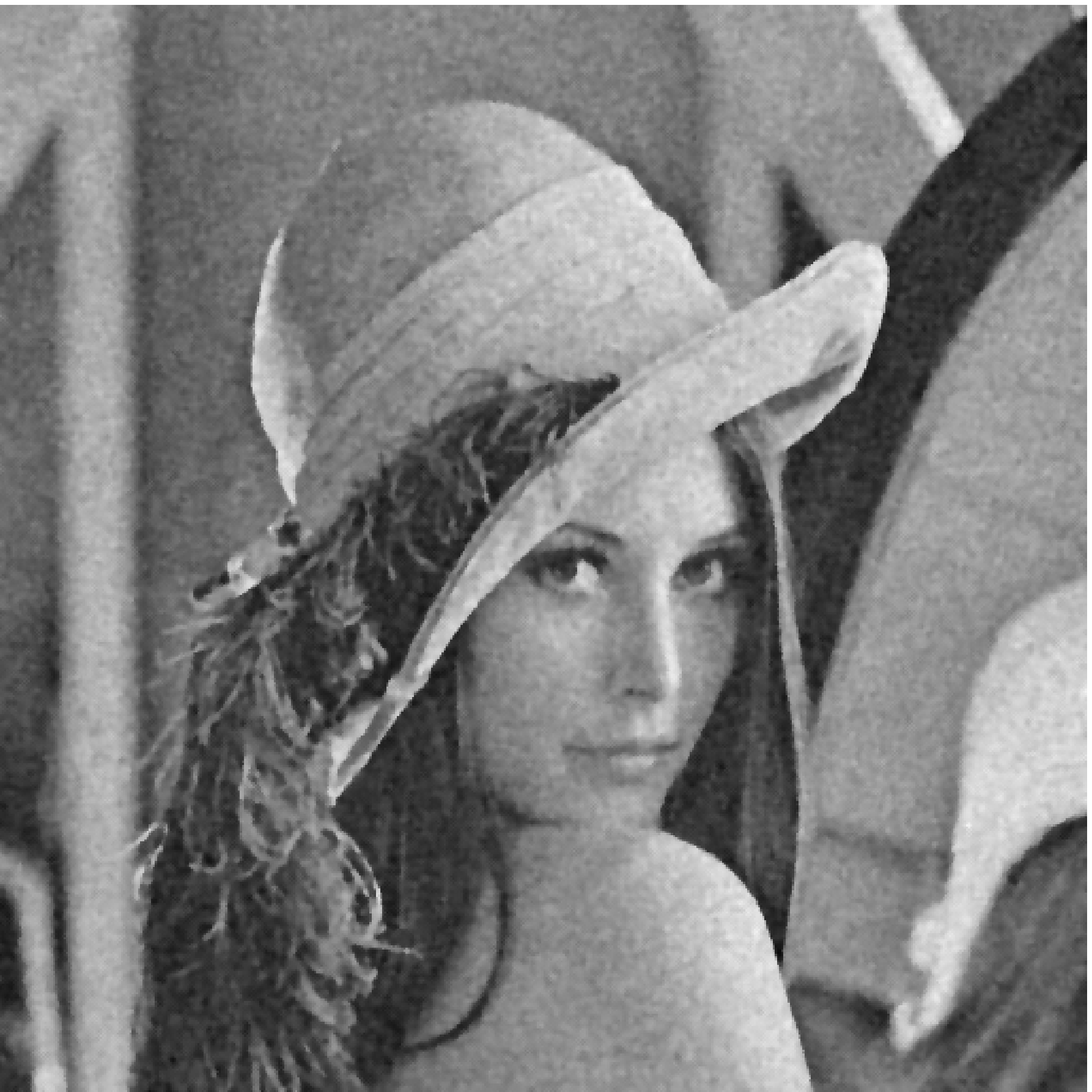} &    \\
\includegraphics[width=0.30\linewidth]{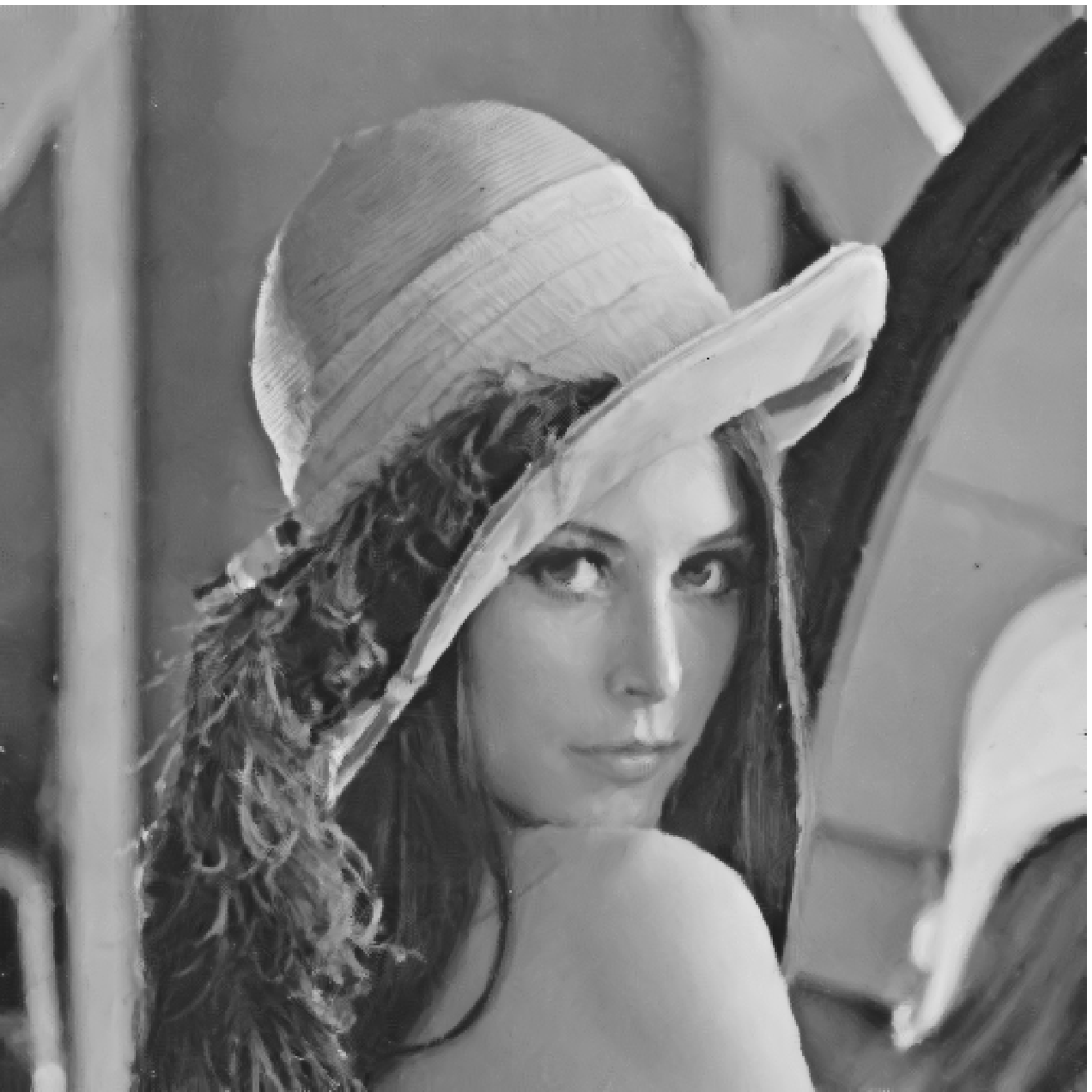} & %
\includegraphics[width=0.30\linewidth]{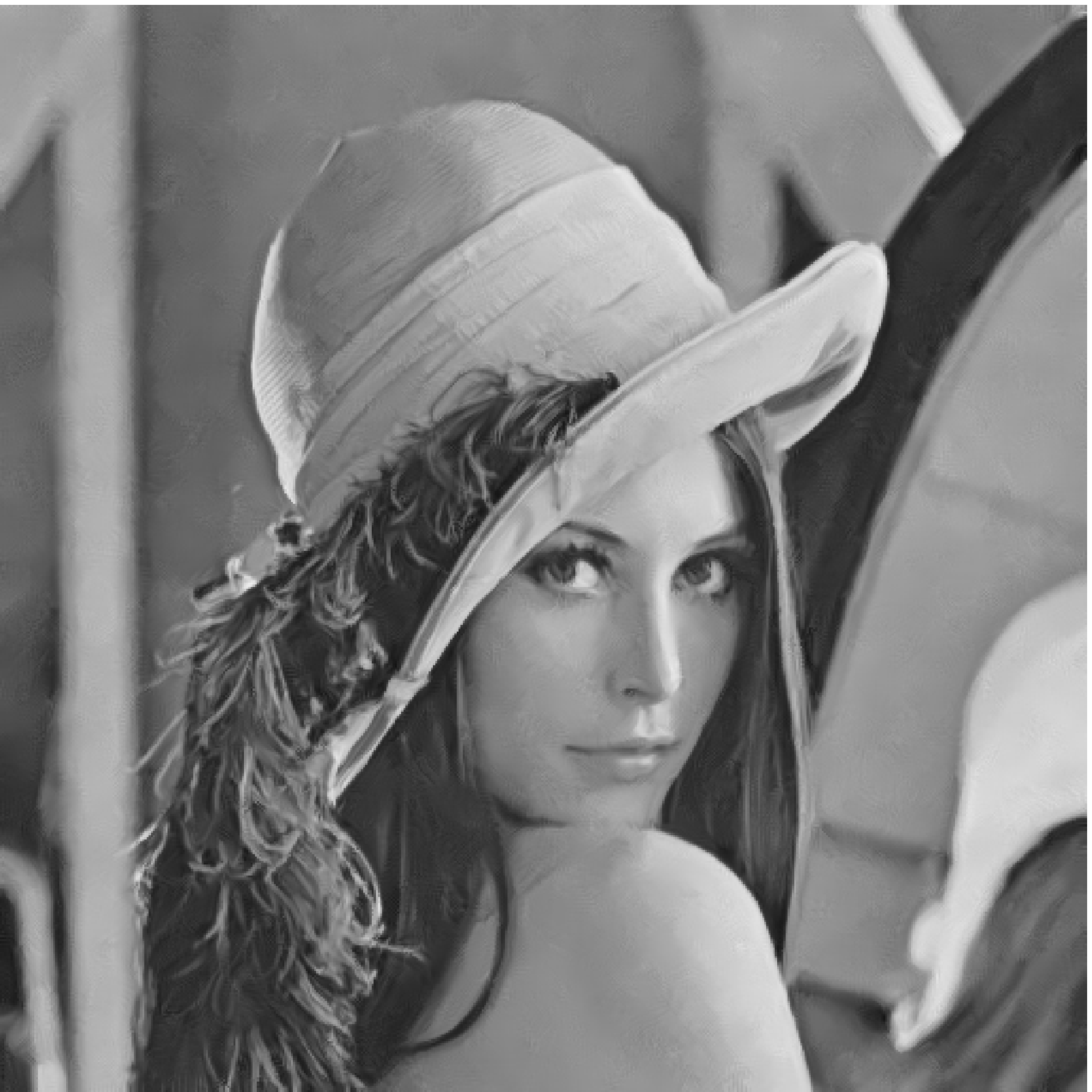} & %
\includegraphics[width=0.30\linewidth]{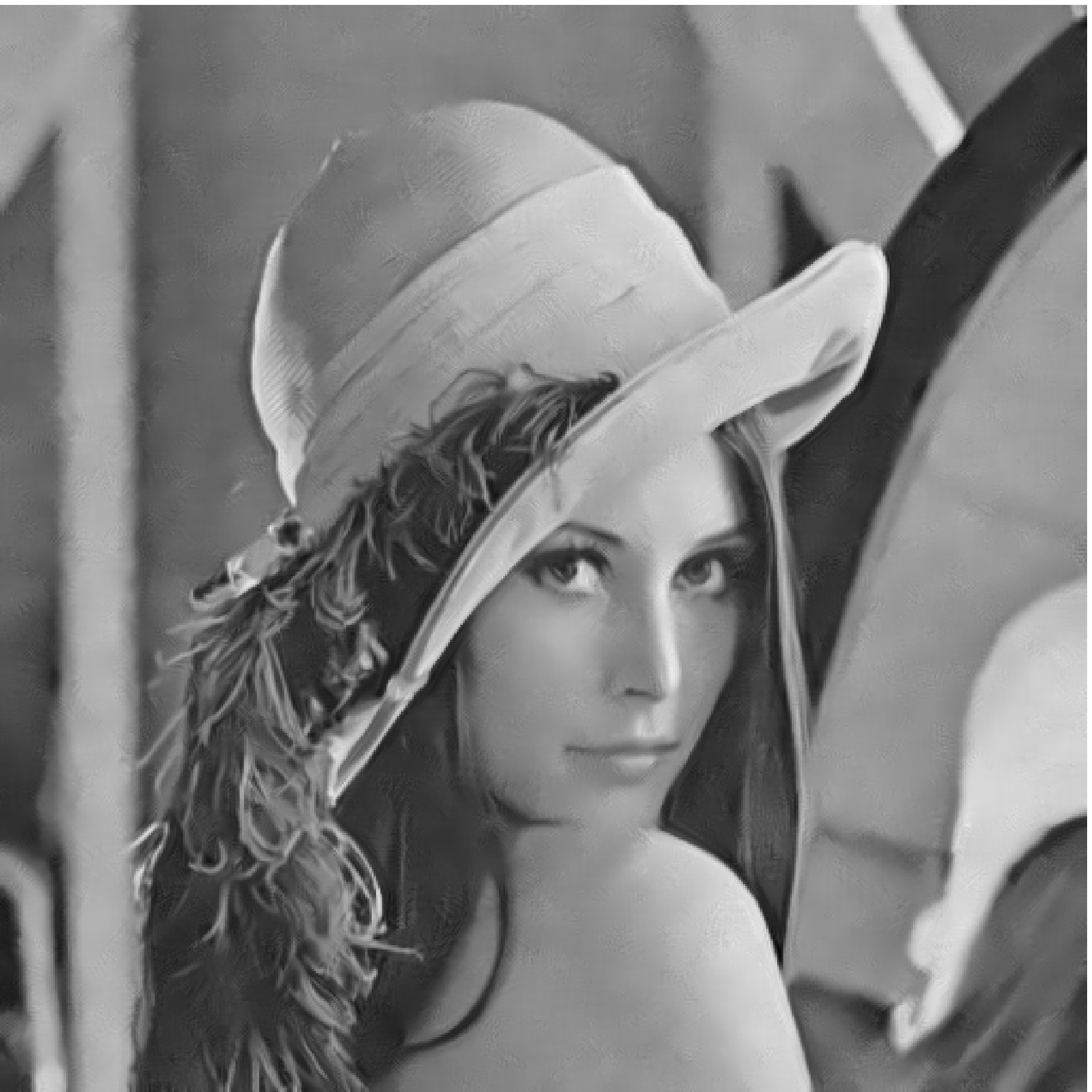} &    \\
\includegraphics[width=0.30\linewidth]{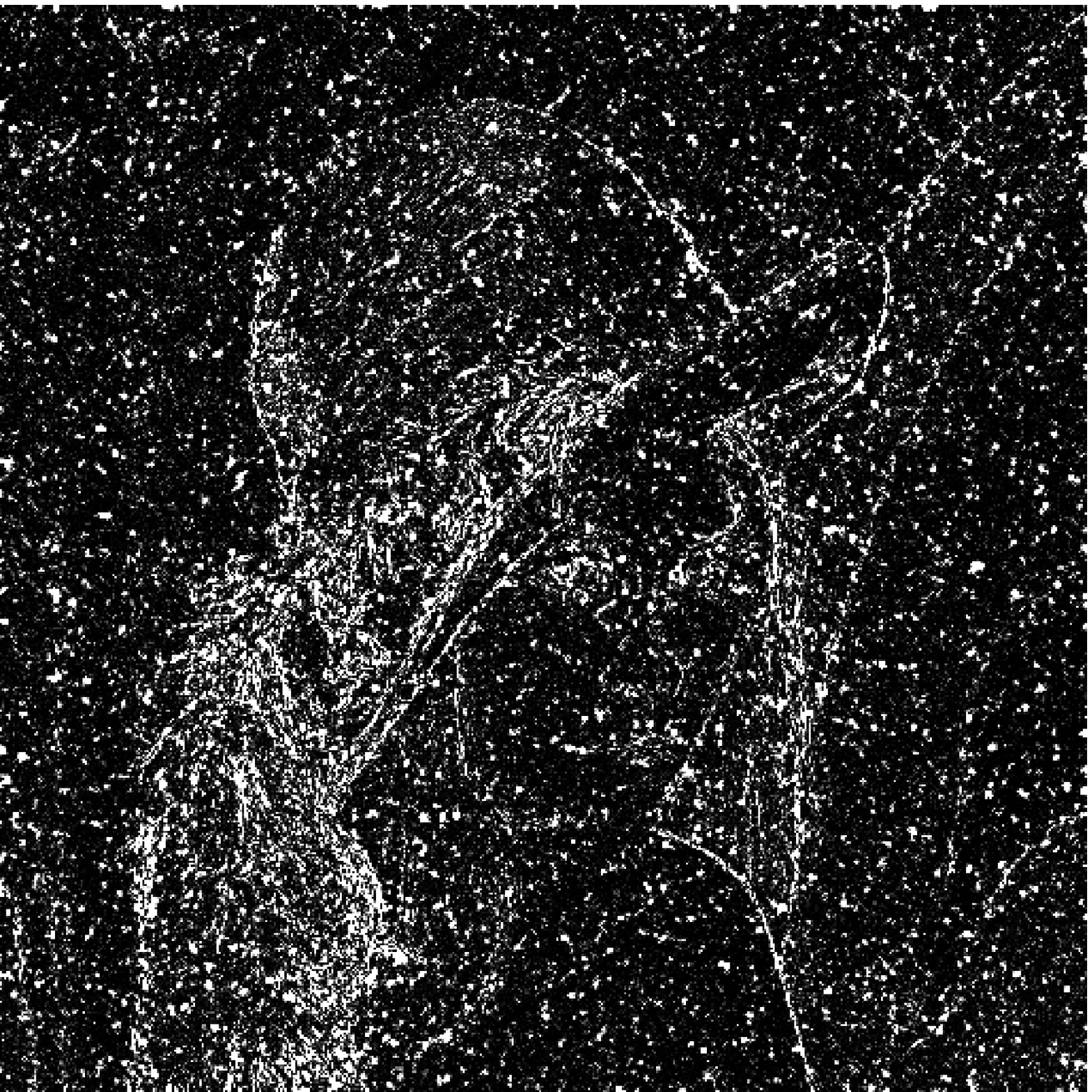} & %
\includegraphics[width=0.30\linewidth]{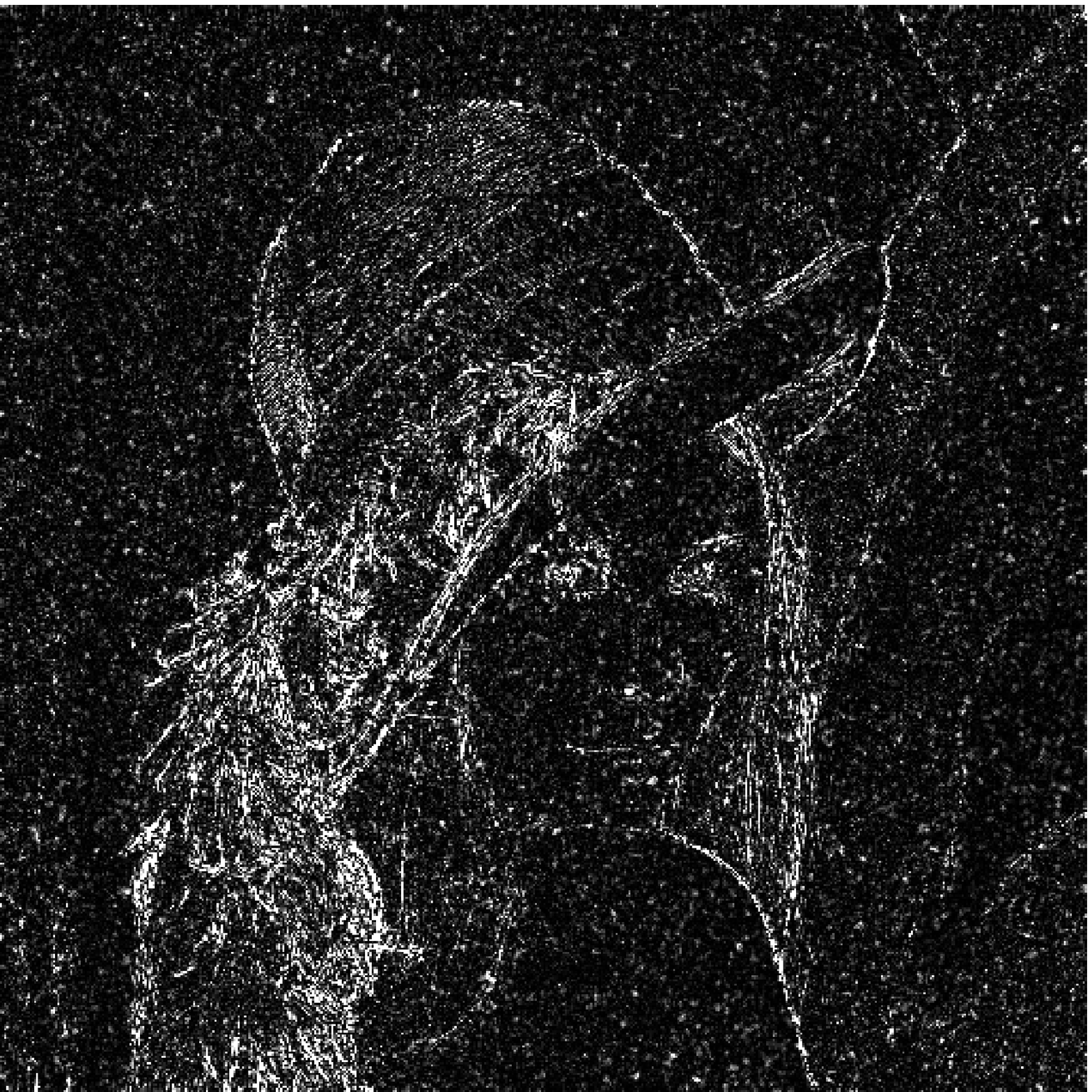} & %
\includegraphics[width=0.30\linewidth]{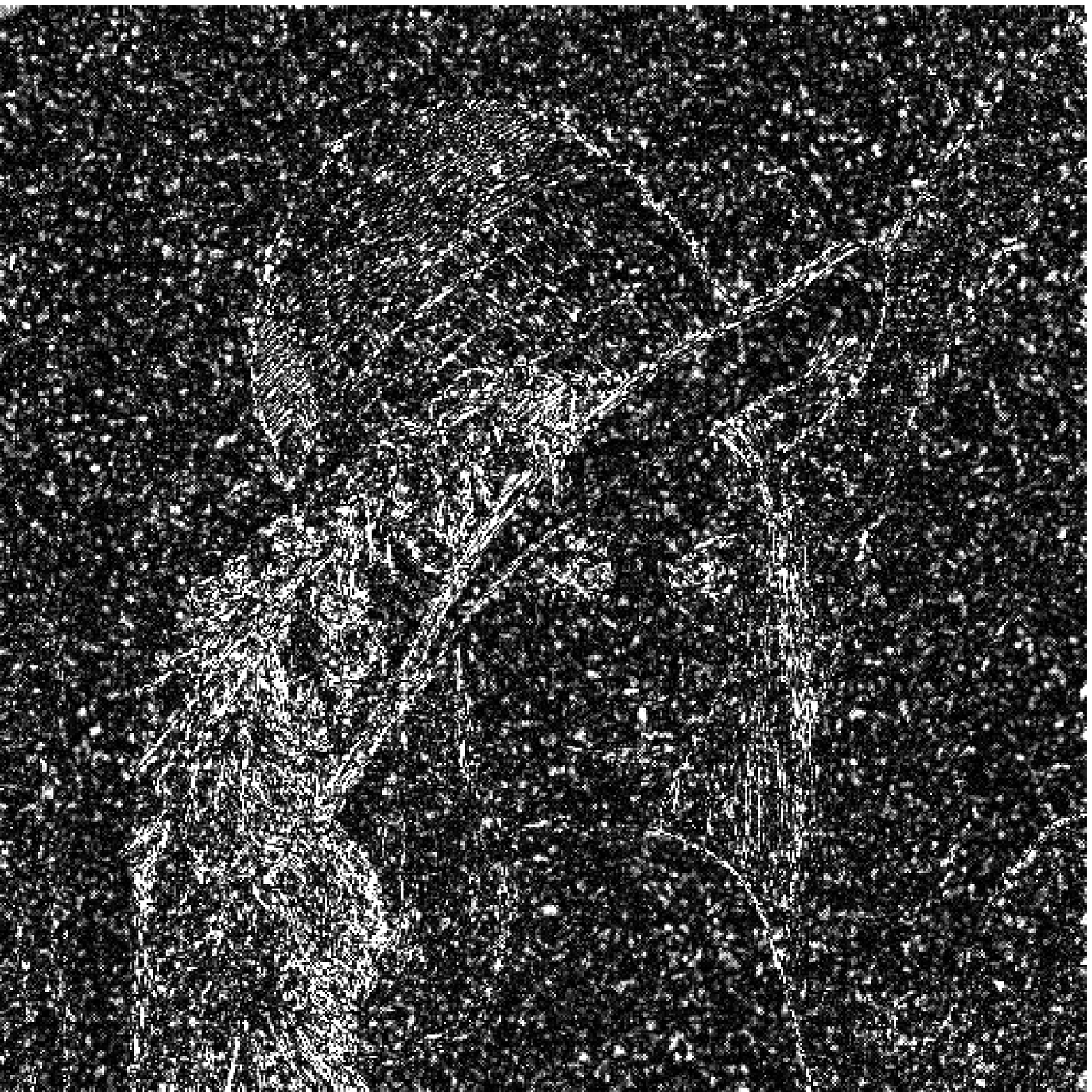} &    \\
\includegraphics[width=0.30\linewidth]{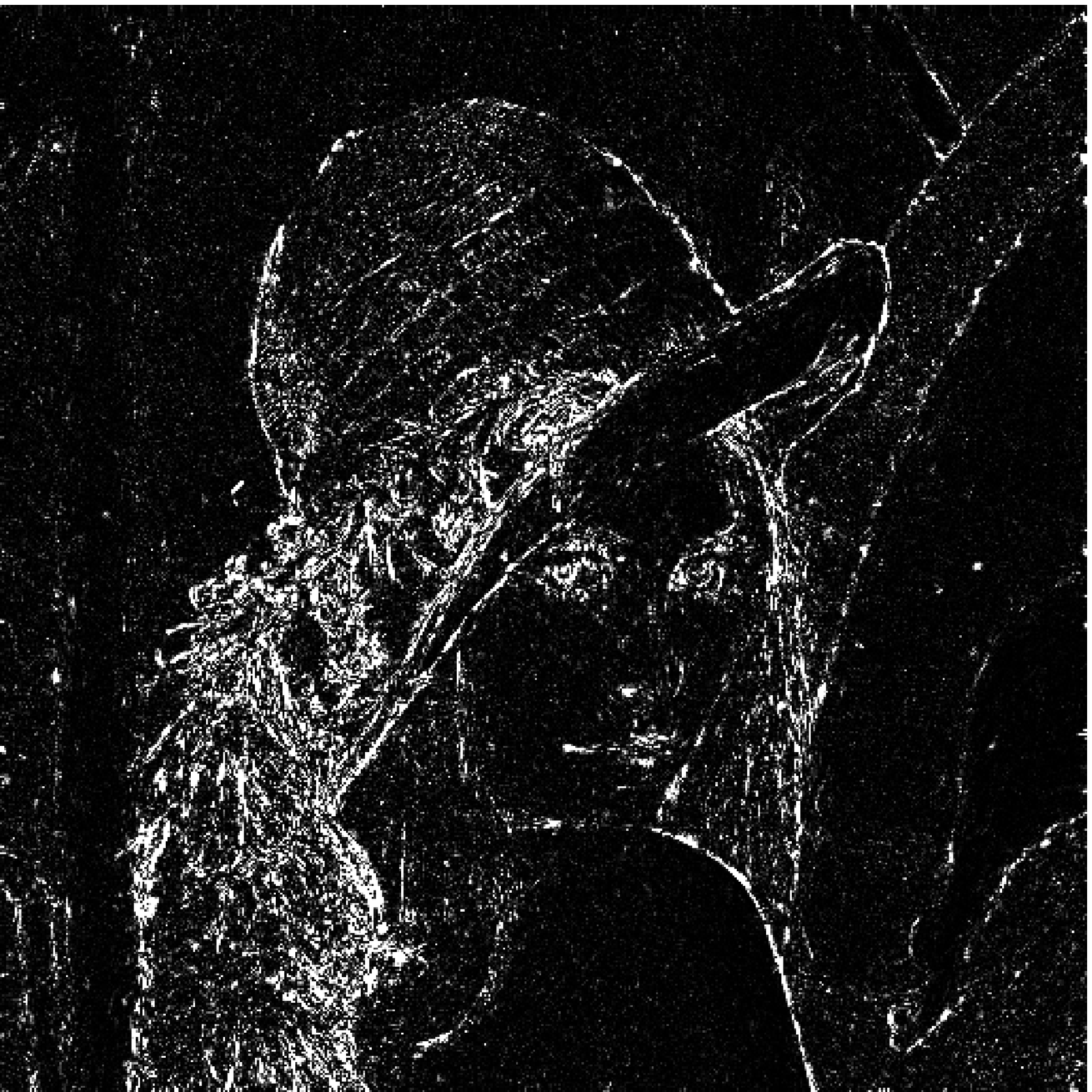} & %
\includegraphics[width=0.30\linewidth]{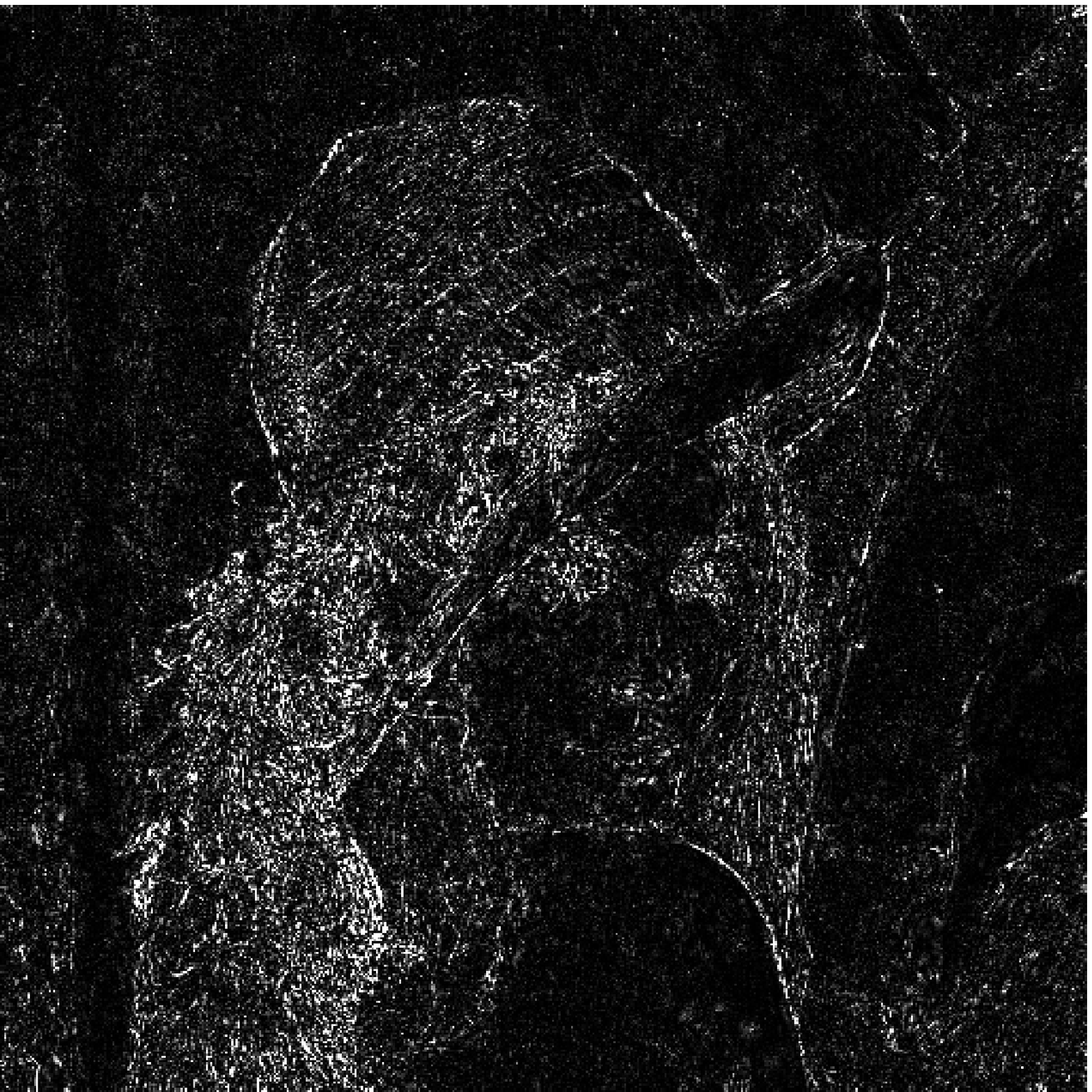} & %
\includegraphics[width=0.30\linewidth]{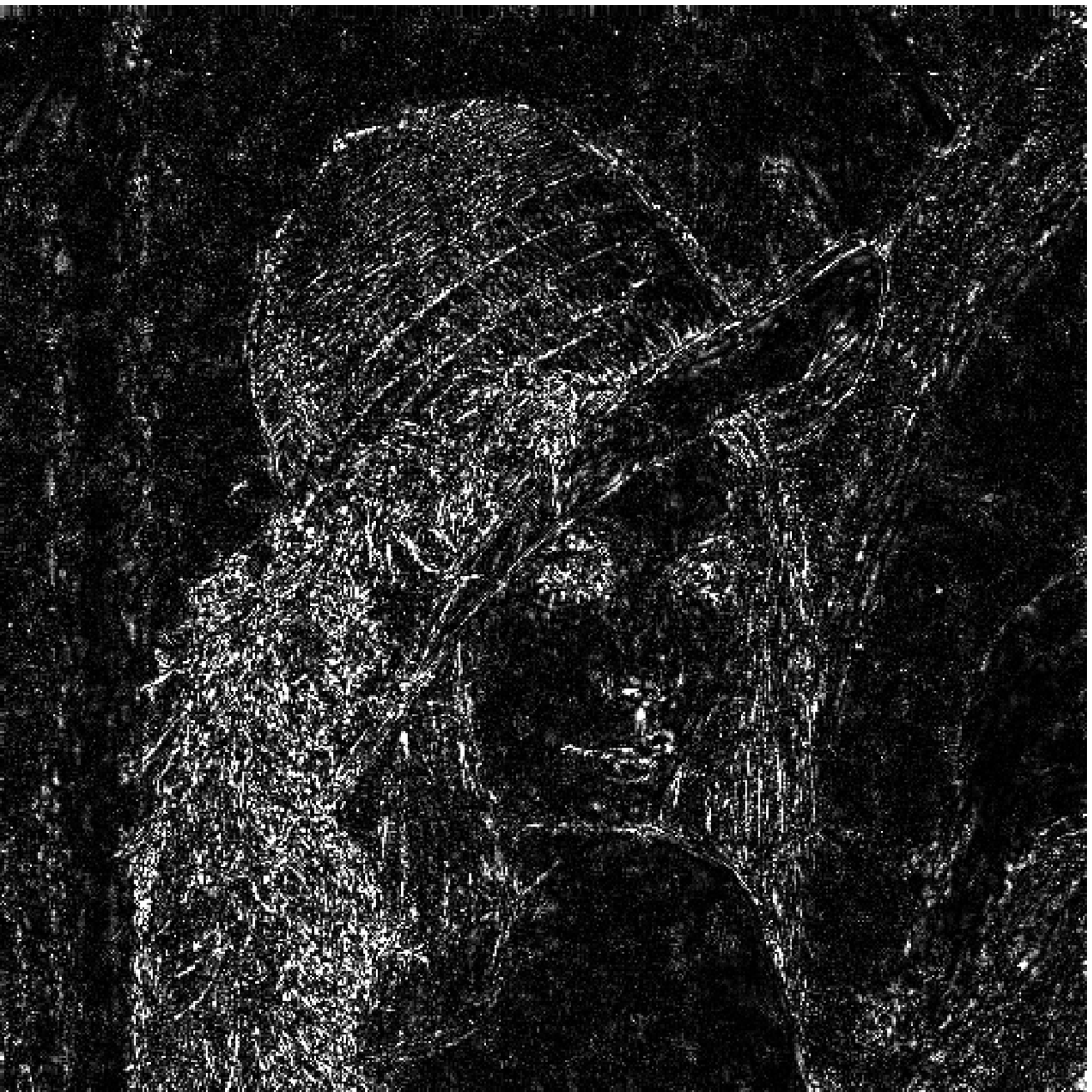} &
\end{tabular}
} \vskip1mm
\par
\rule{0pt}{-0.2pt}%
\par
\vskip1mm 
\end{center}
\caption{{\protect\small The first row is the images with different levels
of mixture of impulse and Gaussian noise. The second and the third row
display the images restored by TriF and OWMF respectively. The forth and the
fifth row give square error image with TriF and OWMF respectively.}}
\label{Figure visual quality}
\end{figure}

 We use the
 kernel $\kappa_{0}$ for computing the estimated brightness variation
function $\widehat{\rho }_{J,\kappa,x_{0}},$ which corresponds to the Optimal
Weights Mixed Filter as defined in Section \ref{sec Optimal Weights
Mixed Filter}. The parameters $m$ and $M$ have been fixed to $m=25\times 25$
and $M=13\times 13.$ In Figure\ \ref{Figure visual quality}, the images in
the third row show that the noise is reduced in a natural manner and
significant geometric features, fine textures, and original contrasts are
visually well recovered with no undesirable artifacts. To better appreciate
the accuracy of the restoration process, the images of square errors (the square of
the difference between the original image and the recovered image) are shown
in the fifth row of Figure\ \ref{Figure visual quality}, where the dark
values correspond to a high-confidence estimate. As expected, pixels with a
low level of confidence are located in the neighborhood of image
discontinuities. For comparison, we show the images denoised by the
trilateral filter TriF (see the images in the second row of Figure
\thinspace\ \ref{Figure visual quality} ) and their square errors (see the
images in the forth row of Figure \thinspace\ \ref{Figure visual quality}).
We can see clearly that the images of square errors of our methods are darker than that of Trif, so our method provides significant improvement over.
The overall visual impression and the numerical results are improved using
our algorithm.

For comparison, we consider  follows three cases: pure Gaussian noise, pure impulse noise and the mixture of Gaussian and impulse noises. In the case of pure Gaussian white noise, we have done simulation on a commonly-used set of images ("Lena", "Barbara", "Boat" and "House") available at http://decsai.ugr.es/~javier/\\
denoise/test\_images/ and the comparison with several filters is given in Table \ref{Table gaussian}.   The  PSNR values show that our approach work as well as those sophisticated methods, like
 \citet{hirakawa2006image},
 \citet{kervrann2008local},
\citet{hammond2008image} and
 \citet{aharon2006rm},
and is better than  the filters proposed in
\citet{buades2005non},
\citet{salmon2009nl},
 \citet{katkovnik2004directional},
\citet{foinovel} and
 \citet{roth2009fields}.  The proposed approach  gives a quality of denoising which is competitive with that of the best recent method BM3D (see \citet{dabov2007image}).
These methods can only deal wits pure Gaussian noise, while our method can not only cope with the Gaussian noise, but also remove the impulse noise and the mixture of Gaussian and pure impulse noises. For the pure impulse noise, our method is also competitive to the sophisticated method. In order to compare the others methods, we choose a commonly set of images ("Baboon", "Badge", "Lena" and "Pentagon") which is taken in \citep{dong2007detection}. Table \ref{Table impulse} lists the restoration results of well-know different algorithm. It is clear  that our method provides significant improvement over
\citet{sun1994detail},
 \citet{abreu1996new},
 \citet{wang1999progressive},
 \citet{chen1999tri},
\citet{chen2001space},
\citet{chen2001adaptive},
 \citet{crnojevic2004advanced},
\citet{wenbin2005new}, etc. Our approach work as well as
 \citet{dong2007detection}
 and  \citet{yu2008efficient}, when our approach produces the best PSNR values in the cases of "Baboon" (40\%) and "Pentagon" (20\% and 40\%), while
 \citet{yu2008efficient} has the best  results in the case of "Baboon" (20\%) and "Bridge" (20\% and 40\%), and  \citet{dong2007detection}(ROLD-EPR) wins in the case of "Lena" (20\% and 40\%).
Finally, we show the comparison between the  \citet{garnett2005universal} and our method with the set of images ("Lena", "Bridge", "Boat" and "Barbara"). From Table \ref{Table mixed}, it is clear that our method provides significant improvement over the  algorithm Trif  \citep{garnett2005universal}.

\section{Conclusion\label{sec conclusion}}

A new image denoising filter, which deal with the mixture of Gaussian and
impulse noises model based on weights optimization and the modified
Rank-Ordered Absolute Differences statistic, is proposed. The implementation of the filter is straightforward. Our work
leads to the following conclusions.

\begin{enumerate}
\item The modified Rank-Ordered Absolute Differences statistic, used in the
new filter, detects effectively the impulse noise in the case of mixture of
Gaussian and impulse noises. This statistic is well adapted for use with the
Weights Optimization Filter of \citep{jin2011removing}.

\item The proposed filter is proven by simulations to be very efficient for
removing both a mixture of impulse and Gaussian noises, and the pure
impulse or pure Gaussian noise.

\item Our numerical results demonstrate that the new filter outperforms the
known filters.
\end{enumerate}
%


\end{document}